\documentclass[preprint,12pt]{elsarticle}

\usepackage{amssymb}
\usepackage{amsmath}

\usepackage{algorithm}
\usepackage{algorithmic}

\usepackage{subfig}
\usepackage{multirow}
\usepackage{xcolor}
\usepackage{amsmath}
\usepackage{booktabs}
\usepackage[hyphens]{url}
\usepackage{hyperref}
\hypersetup{colorlinks=true,breaklinks=true}

\journal{Information Fusion}

\begin{document}

\begin{frontmatter}



\title{HDRT: A Large-Scale Dataset for Infrared-Guided HDR Imaging}

\author[label1,label2]{Jingchao Peng}
\author[label1]{Thomas Bashford-Rogers}
\author[label3]{Francesco Banterle}
\author[label2]{Haitao Zhao \corref{cor1}}
\cortext[cor1]{Corresponding author.}
\author[label1]{Kurt Debattista}

\affiliation[label1]{organization={Warwick Manufacturing Group, University of Warwick},
            city={Coventry},
            postcode={CV47AL},
            country={UK}}

\affiliation[label2]{organization={School of Information Science and Engineering, East China University of Science and Technology},
            addressline={Meilong Road 130},
            city={Shanghai},
            postcode={200237},
            country={China}}

\affiliation[label3]{organization={Institute of Information Science and Technologies, Consiglio Nazionale delle Richerch},
            addressline={Via Giuseppe Moruzzi 1},
            city={Pisa},
            postcode={56124},
            country={Italy}}

\begin{abstract}
Capturing images with enough details to solve imaging tasks is a long-standing challenge in imaging, particularly due to the limitations of standard dynamic range (SDR) images which often lose details in underexposed or overexposed regions. Traditional high dynamic range (HDR) methods, like multi-exposure fusion or inverse tone mapping, struggle with ghosting and incomplete data reconstruction.
Infrared (IR) imaging offers a unique advantage by being less affected by lighting conditions, providing consistent detail capture regardless of visible light intensity.
In this paper, we introduce the HDRT dataset, the first comprehensive dataset that consists of HDR and thermal IR images.
The HDRT dataset comprises 50,000 images captured across three seasons over six months in eight cities, providing a diverse range of lighting conditions and environmental contexts.
Leveraging this dataset, we propose HDRTNet, a novel deep neural method that fuses IR and SDR content to generate HDR images. Extensive experiments validate HDRTNet against the state-of-the-art, showing substantial quantitative and qualitative quality improvements.
The HDRT dataset not only advances IR-guided HDR imaging but also offers significant potential for broader research in HDR imaging, multi-modal fusion, domain transfer, and beyond.
The dataset is available at https://huggingface.co/datasets/jingchao-peng/HDRTDataset.
\end{abstract}

\begin{graphicalabstract}
\includegraphics[width=1.0\textwidth]{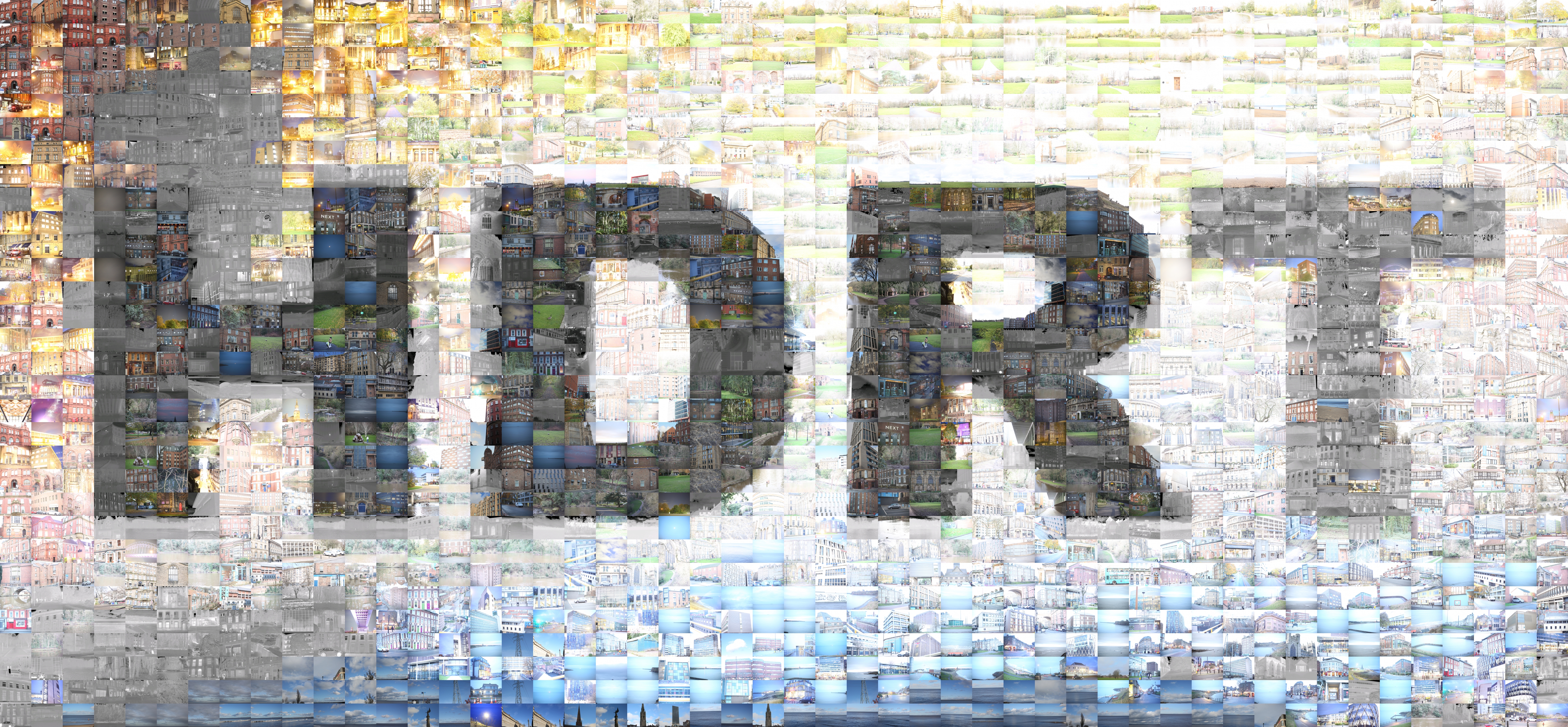}
\vspace{-0.7cm}
\captionof{figure}{We introduce HDRT, a comprehensive dataset including high dynamic range images (10k), standard dynamic range images with different exposure levels (30k), and infrared images (10k) collected from eight cities situated at various latitudes, thereby ensuring a diverse range of lighting conditions and environmental contexts. 
\emph{Zoom in to see examples of the dataset.}}
\label{fig:An-example-of-the-pro}
\end{graphicalabstract}



\end{frontmatter}



\section{Introduction}


Most applications in computer vision and image understanding rely on standard dynamic range 8-bit RGB data \cite{HDR1}. However, a significant body of work \cite{face_maching, fault_detection, HDR_detection, face_detection} has shown that the capture limitations inherent in SDR data lead to suboptimal performance. These limitations result from two aspects: the limited dynamic range of the captures means that details are frequently missing in clipped image regions \cite{Eil, Tel_dataset}, and the restriction to imaging in the visible spectrum means that details that could aid imaging applications \cite{hyperspectral, DFNet} are not acquired.

High dynamic range (HDR) imaging solves the first limitation and permits the capture, transmission, and visualization of real-world lighting \cite{HDR2, HDR3, HDR4}. Typically, HDR image capture relies on multiple exposure fusion methods \cite{Debevec+1997,Lecouat+2022}. However, due to the necessity of capturing multiple images simultaneously, these multiple exposure fusion methods can often introduce ghosting artifacts during the merging or limit the number of exposures; this means they are capable of capturing only a medium dynamic range.

Imaging outside the visible spectrum solves the second limitation by using information from the spectrum of wavelengths outside the visible domain. We focus on infrared (IR) imaging as IR typically serves as a supplement to visible light imagery under poor visual conditions, thereby providing information in cases of underexposure. This has been exploited in multiple applications from object tracking \cite{SiameseTracking} to low-light enhancement \cite{RGBTLowLight}. Moreover, since infrared imaging essentially measures the temperature of an object, excessive visible illumination often only marginally affects infrared imaging, enabling the acquisition of information in cases of overexposed images as well.
This capability has led to a growing trend of combining visible light and IR modalities, both in industry and academia. In industry, the integration of thermal and visible light imaging has become increasingly popular, particularly in consumer electronics. For instance, the AGM Glory G1S includes synchronized visible light and IR cameras; the FLIR ONE series devices can be used with smartphones to capture IR images on mobile devices.

These two observations highlight the need for imaging outside the standard SDR RGB range, and their combination will further enable new applications in imaging. However, this research needs to be enabled by a large, robust dataset which includes paired HDR and IR data. 
While there are existing datasets \cite{Kalantari+2017, Mobile_HDR, LasHeR, FLIR} that support either HDR or IR imaging, none provide both modalities together.
This paper is the first to propose such a dataset; see Fig. \ref{fig:An-example-of-the-pro}.

Leveraging this extensive dataset, we propose a novel deep-learning-based method, HDRTNet, which fuses IR with visible single-exposure RGB data to generate HDR images, thereby providing a new approach to HDR image capture. A desirable HDR image capture method should require minimal capture time, and should contain details in both under and over exposed regions. While this seems to be a contradictory problem in the visible domain, we make the observation that much of the information required is available outside the visible domain, and this extra information can be used to solve both problems simultaneously. Our dataset therefore enables this type of application. Extensive experiments show that our proposed method can successfully fuse information from IR and visible domains to generate HDR images, especially under extreme visual conditions. We perform quantitive and qualitative comparisons against the state of the art and show that our method outperforms, often substantially, conventional single-image capture methods.

The proposed HDRT dataset is not only valuable for IR-guided HDR imaging but also serves as a significant resource for broader HDR research, including multi-exposure fusion and inverse tone mapping. Additionally, it is a valuable asset for multimedia tasks, such as RGBT fusion and domain transfer, and supports various other visual perception and image reconstruction endeavors. To summarize, the main contributions of this paper are: 

\begin{itemize}
\item We propose the first aligned IR and HDR dataset in the world, consisting of 40K HDR and SDR images and 10K IR images, to facilitate research on RGBT-fusion-based HDR imaging and other multimodal research. 
\item We show a practical capture approach which can rapidly generate such RGBT datasets, and discuss solutions to technical challenges during the capture and post-process steps.
\item To the best of our knowledge, we are the first to introduce the use of the IR domain to enhance HDR imaging.
\item We develop HDRTNet, a novel and practical deep learning approach which fuses RGB and IR imagery to generate HDR images.
%
\item Results show that HDRTNet achieves better performance using IR images and single-exposed SDR images than previous methods, and can successfully resolve the missing data problem in single image captures.
%

\end{itemize}

\section{Related Work}
\label{sec:relatedwork}
HDR content can be generated using a number of methods, such as inverse tone mapping and multiple exposure methods. 
In this section, we will present all these approaches and provide a primer on RGBT fusion, alongside a discussion of related HDR and RGBT datasets, as summarized in Tab. \ref{tab:dataset-comparison}.\\


\begin{table*}[!t]
\caption{\label{tab:dataset-comparison} Comparison between related datasets and the proposed HDRT dataset.}
\centering{}\resizebox{\textwidth}{!}{
\begin{tabular}{lcccccl}
\hline
\toprule
Dataset     & Total Frames & Number of Scenes & Resolution                                                                                       & HDR Images & IR Images & Description                                  \\
\midrule
GTOT \cite{GTOT}        & 15,800       & 50               & 384\texttimes{}288                                                                                          & \texttimes{}               & \textsurd{}              & SDR and IR video sequence for RGBT tracking. \\
RGBT234 \cite{RGBT234}     & 234,000      & 234              & 630\texttimes{}460                                                                                          & \texttimes{}               & \textsurd{}              & SDR and IR video sequence for RGBT tracking. \\
LasHeR  \cite{LasHeR}      & 734,800      & 1,224             & 630\texttimes{}480                                                                                          & \texttimes{}               & \textsurd{}              & SDR and IR video sequence for RGBT tracking. \\
LLVIP \cite{LLVIP}     & 30,976      & 15,488             & 1040\texttimes{}720                                                                                          & \texttimes{}               & \textsurd{}              & SDR and IR  image pairs for low-light vision. \\
FLIR ADAS \cite{FLIR}     & 26,442      & 7,498             & 640\texttimes{}512                                                                                          & \texttimes{}               & \textsurd{}              & SDR and IR  image pairs for ADAS. \\
Kalantari \cite{Kalantari+2017}  & 356         &  89              & 1500\texttimes{}1000                                                                                        & \textsurd{}               & \texttimes{}              & HDR and SDR images for HDR imaging.          \\
NTIRE 2021 \cite{NTIRE21}  & 6,216        & 1,554            & 1900\texttimes{}1060                                                                                        & \textsurd{}               & \texttimes{}              & HDR and SDR images for HDR imaging.          \\
Tel \cite{Tel_dataset}        & 576          & 144              & 1496\texttimes{}1000                                                                                        & \textsurd{}               & \texttimes{}              & HDR and SDR images for HDR imaging.    
\\
Mobile-HDR \cite{Mobile_HDR}        & 1,124          & 281              & 4608\texttimes{}3456                                                                                        & \textsurd{}               & \texttimes{}              & HDR and SDR images captured by mobile phone.    \\ \midrule
HDRT (Ours) & 50,000       & 10,000           & \begin{tabular}[c]{@{}c@{}}5120\texttimes{}3840 (RGB)\\ 1280\texttimes{}960 (IR)\end{tabular} & \textsurd{}               & \textsurd{}              & HDR, SDR, and IR images for HDR imaging.   \\ 
\bottomrule
\end{tabular}}
\end{table*}

\noindent\textbf{HDR Methods and Datasets.}
Multiple exposure HDR generation requires the capture of the same scene at different exposure times to create HDR images \cite{Debevec+1997, Lecouat+2022}.
However, these can produce motion artifacts and alignment issues; they can increase the complexity of the workflow and be time-consuming. 
Inverse tone mapping (ITM) proposes to extend the dynamic range of SDR content using a single image as input.
This is an ill-posed problem, and different solutions have been proposed.
Recently, the use of an end-to-end approach has improved the reconstruction by leveraging the superior ability to model complex, non-linear mappings of deep-learning architectures \cite{Eil, yu2021luminance, Zhang+2021}.
Other approaches of generating lower and higher exposure images have been employed in many works using GAN during training to improve the final quality \cite{DeepRecursiveHDRI} or via transformers to determine only necessary exposures \cite{Zhang_2023_CVPR}.

To support these deep-learning methods, several HDR datasets have been developed.
Kalantari et al. \cite{Kalantari+2017} introduced a dataset consisting of three SDR input images of the same scene with different exposures, pioneering HDR datasets for training multiple exposure methods.
Tel et al. \cite{Tel_dataset} expanded on this by capturing HDR images under diverse lighting conditions, though both datasets contain fewer than 150 scenes.
The NTIRE HDR challenge \cite{NTIRE21} introduced a more diverse dataset by capturing HDR videos with professional rigs and generating corresponding SDR images synthetically; however, the synthetic SDR images often contain non-faithful inputs due to noise.
Prabhakar et al. \cite{Prabhakar_dataset} presented a dataset claiming over 500 samples, but only 32 are publicly available, limiting its utility.
Due to the lack of large datasets of HDR content, researchers have started to investigate the use of SDR datasets for static images \cite{wang2022glowgan} and self-supervision on input SDR videos \cite{Banterle+2024}. \\

\noindent\textbf{RGBT Fusion and Multimodal Datasets.}
Visible light and infrared image fusion, known as RGBT fusion \cite{Corneanu+2016}, which integrates visible light and infrared (IR) imagery, has garnered significant attention for its ability to enhance performance under challenging lighting conditions \cite{RGBTFusion, RGBTTracking}.
RGBT fusion integrates thermal information with visible-light-based methods to enhance performance under degraded lighting conditions and facilitate other task performance such as detection \cite{RGBTdetection}, tracking \cite{DFNet}, and color-vision-related tasks \cite{RGBTColor}.
RGBT fusion strategies are typically categorized into pixel-level, decision-level, and feature-level fusion \cite{RGBTTracking}.
Pixel-level fusion treats the IR modality as an additional channel in the RGB images \cite{Pixel2}, while decision-level fusion separately processes RGB and IR images before combining their outputs \cite{Decision}.
Feature-level fusion, on the other hand, focuses on extracting and integrating features from RGBT image pairs in separate feature spaces through different sub-networks \cite{SiameseTracking}, making it more suitable for HDR tasks by avoiding dependency on individual patterns or strict image registration.

Various datasets were developed to support RGBT fusion research.
The GTOT dataset \cite{GTOT} included 15,800 frames of synchronized RGB and IR video sequences across 50 scenes, and the RGBT234 dataset \cite{RGBT234} provided 234,000 frames of RGB and thermal data for multi-modal tracking.
The LasHeR dataset \cite{LasHeR} offered a larger collection of 734,800 frames across 1,224 scenes.
All those datasets primarily served RGBT tracking but lacked HDR content, limiting their application in HDR image tasks.
The LLVIP dataset \cite{LLVIP} focused on low-light visual perception with 30,976 paired RGB and thermal images.
The FLIR ADAS dataset \cite{FLIR} was designed for automotive applications, providing thousands of paired RGB and thermal images captured from a vehicle perspective.
Despite their contributions, these datasets are limited by their reliance on SDR images, which are insufficient for training models that require high dynamic range inputs.

Although generating infrared data from HDR datasets or creating HDR images from RGBT datasets can partially bridge the gap, deep learning models trained on synthetic data often encounter the risk of model collapse \cite{collapse}.
Therefore, there is a lack of a large-scale dataset that enables the development of more robust models capable of handling complex lighting conditions and integrating multiple modalities, thus leaving a critical gap in the current research landscape.

\section{HDRT Dataset}

We have collected the first HDR and thermal dataset (HDRT) with aligned thermal and both SDR and HDR images in the world. 
This section elaborates on the sensor suite, data processing, and statistics.
\emph{We will make this dataset publicly available upon acceptance}.

\begin{table*}[!t]
\caption{\label{tab:HDRT-dataset-stat} HDRT dataset statistics.}
\centering{}\resizebox{\textwidth}{!}{
\begin{tabular}{lccccccccc}
\toprule
Features & Daytime & Nighttime    & Buildings & Churches & Skyscrapers     & Castles & Bridges & Pedestrians & Boats  \\
Number   & 7571    & 2429         & 8660      & 583      & 1289            & 222     & 452     & 641         & 218    \\ \midrule
Features & Cars    & Water Bodies & Urban     & Rural    & Natural Scenery & Trees   & Parks   & Sky         & Clouds \\
Number   & 689     & 2008         & 4711      & 2191     & 3098            & 4871    & 937     & 1646        & 4669   \\ \bottomrule
\end{tabular}}
\end{table*}

\subsection{Image Capturing}

\begin{figure}[!t]
\centering
\includegraphics[width=0.7\textwidth]{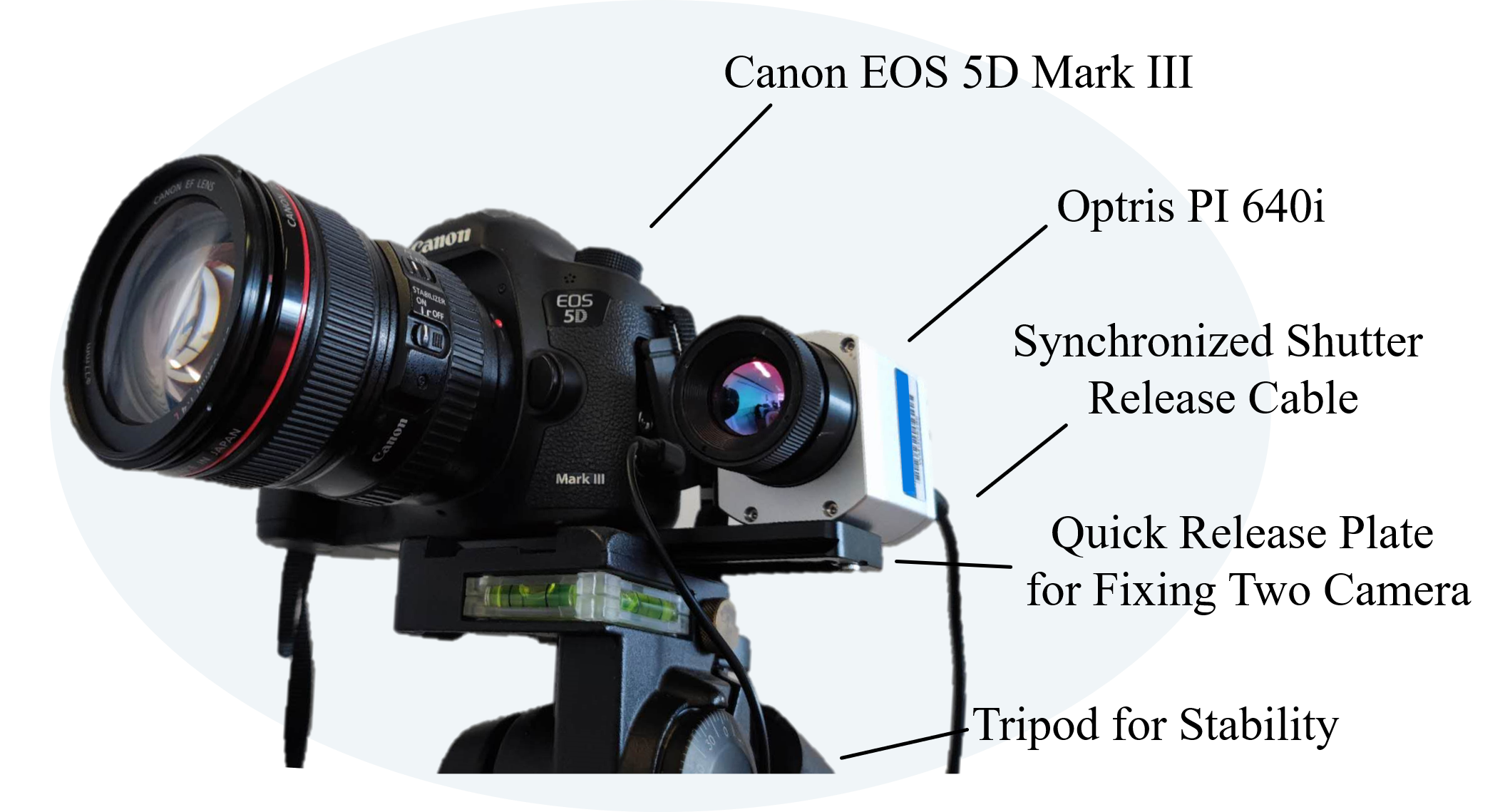} 
\caption{\label{fig:The-capture-hardware}The capture hardware.}
\end{figure}

For capturing IR images, we used an Optris PI 640i camera with 640$\times$480 pixel resolution and working wavelengths from 8 to 14 $\mu$m.
This captures a temperature range from -20 $^\circ$C to 100 $^\circ$C.
We employed EDSR \cite{EDSR} to enhance the resolution to 1280$\times$960. The public dataset provides both resolutions.
We employed a Canon EOS 5D Mark III camera to collect the SDR images.
For each scene captured, we synchronized the shutter release of the infrared camera and the RGB camera. 
The infrared camera collects a single infrared image.
The visible light camera, operating in high-speed continuous shooting mode, captures three SDR images with different exposure values. 
These SDR images are later processed to reconstruct HDR images.
During the capturing process, the RGB and IR cameras are mounted on the same tripod to minimize the registration error due to jitter.
The hardware settings can be shown in Fig \ref{fig:The-capture-hardware}.

\begin{figure}[!t]
\subfloat[Single registration parameter is not suitable for multiple scenes.]{\includegraphics[width=0.48\columnwidth]{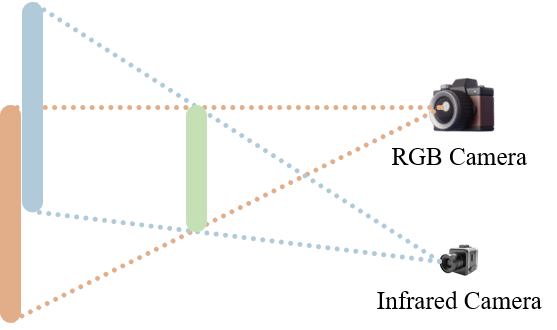}}
\hspace{0.03\columnwidth}
\subfloat[Illustration of the manual registration process.]{\includegraphics[width=0.48\columnwidth]{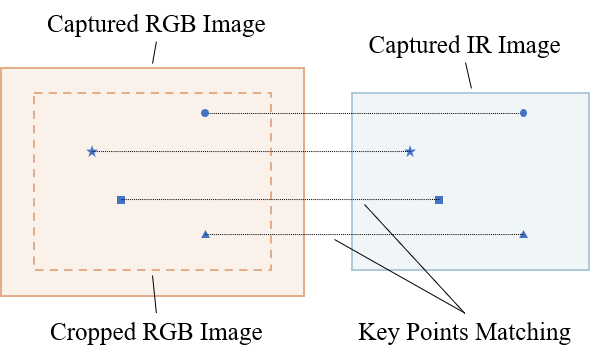}}
\caption{\label{fig:manual-registration}One single registration parameter does not work for all scenes. For example, in (a), when green objects align in one scene, pink and blue objects in another will be misaligned. Therefore, we manually register RGB and IR images as (b). }
\end{figure}

\subsection{Data Processing}

Since visible light images and infrared images are captured by different cameras, binocular parallax is inevitably introduced.
Moreover, the dataset comprises 10,000 scenes, and the diversity of these scenes results in varying shooting distances. 
One homography matrix estimated from a single scene cannot ensure the alignment of all visible light images and infrared images, as seen in Fig. \ref{fig:manual-registration} (a).
Therefore, after capturing, we manually register images from two cameras to ensure accurate alignment between the thermal and visible images.
During the registration process, distinct key points are manually identified and matched in both the visible light and infrared images. 
Due to the higher resolution of the visible light camera compared to the infrared camera, we ensured that the field of view of the infrared camera was contained within that of the visible light camera during capture. 
So, during processing, we only cropped the visible light images, thereby preserving the whole infrared image.
After that, we extracted the overlapping regions to create standard rectangular images.
The illustration of the manual registration process can be seen in Fig. \ref{fig:manual-registration} (b).
Finally, cropped SDR images with different exposure values are merged to generate HDR images by the method of Debevec et al. \cite{Debevec+1997}.

\begin{figure}[!t]
\includegraphics[width=1.0\columnwidth]{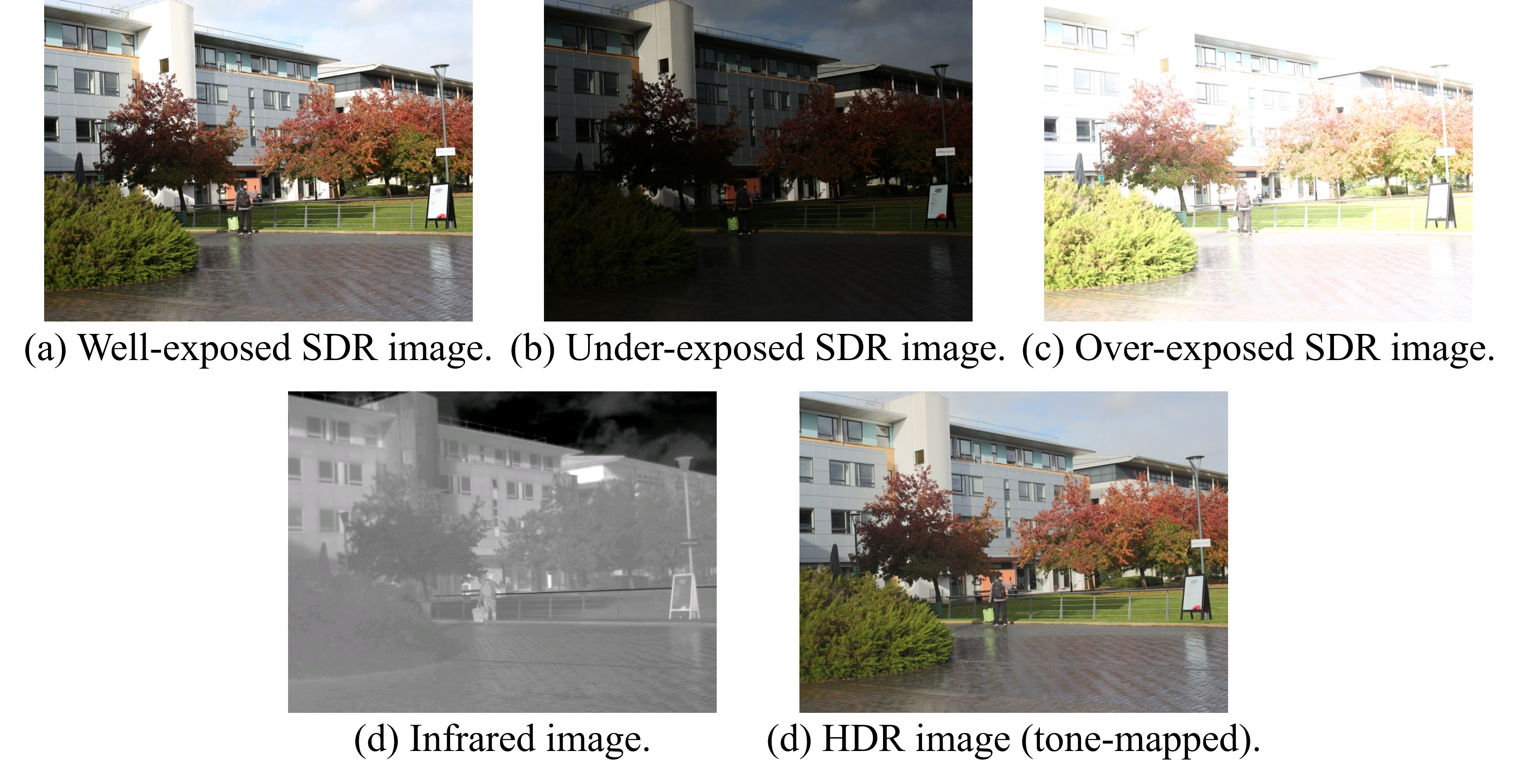}
\caption{\label{fig:example-of-one-cap}Example of one capturing scene.}
\end{figure}

\subsection{Dataset Summary}

HDRT dataset is captured meticulously across three distinct seasons over six months, which encompasses a total of 50,000 images collected from eight different cities situated at various latitudes, thereby ensuring a diverse range of lighting conditions and environmental contexts.
HDRT comprises 10,000 infrared images with a resolution of 1280\texttimes{}960, 30,000 SDR images, and 10,000 HDR images, both with a high resolution of 5120\texttimes{}3840.
The dataset not only provides a rich source of data for visual computing but also includes densely annotated subtitles, enhancing its utility for tasks requiring detailed image descriptions.
The corresponding statistics are outlined in Tab. \ref{tab:HDRT-dataset-stat}. 
It is worth noting that our dataset comprises various lighting conditions, including underexposed and overexposed images, to investigate the performance under extreme visual conditions.
The images contained in a standard capturing scene are shown in Fig. \ref{fig:example-of-one-cap} (see Fig. \ref{fig:An-example-of-the-pro} for a more intuitive representation). 


\section{Experimental Results}

To demonstrate the usefulness of the proposed HDRT dataset, we present a method that makes use of a single exposure and IR to generate an HDR image. We then compare the results against other methods that generated HDR, by evaluating against the ground-truth generated from HDRT. 


\subsection{Methodology}

\begin{figure}[!t]
\centering
\includegraphics[width=0.8\textwidth]{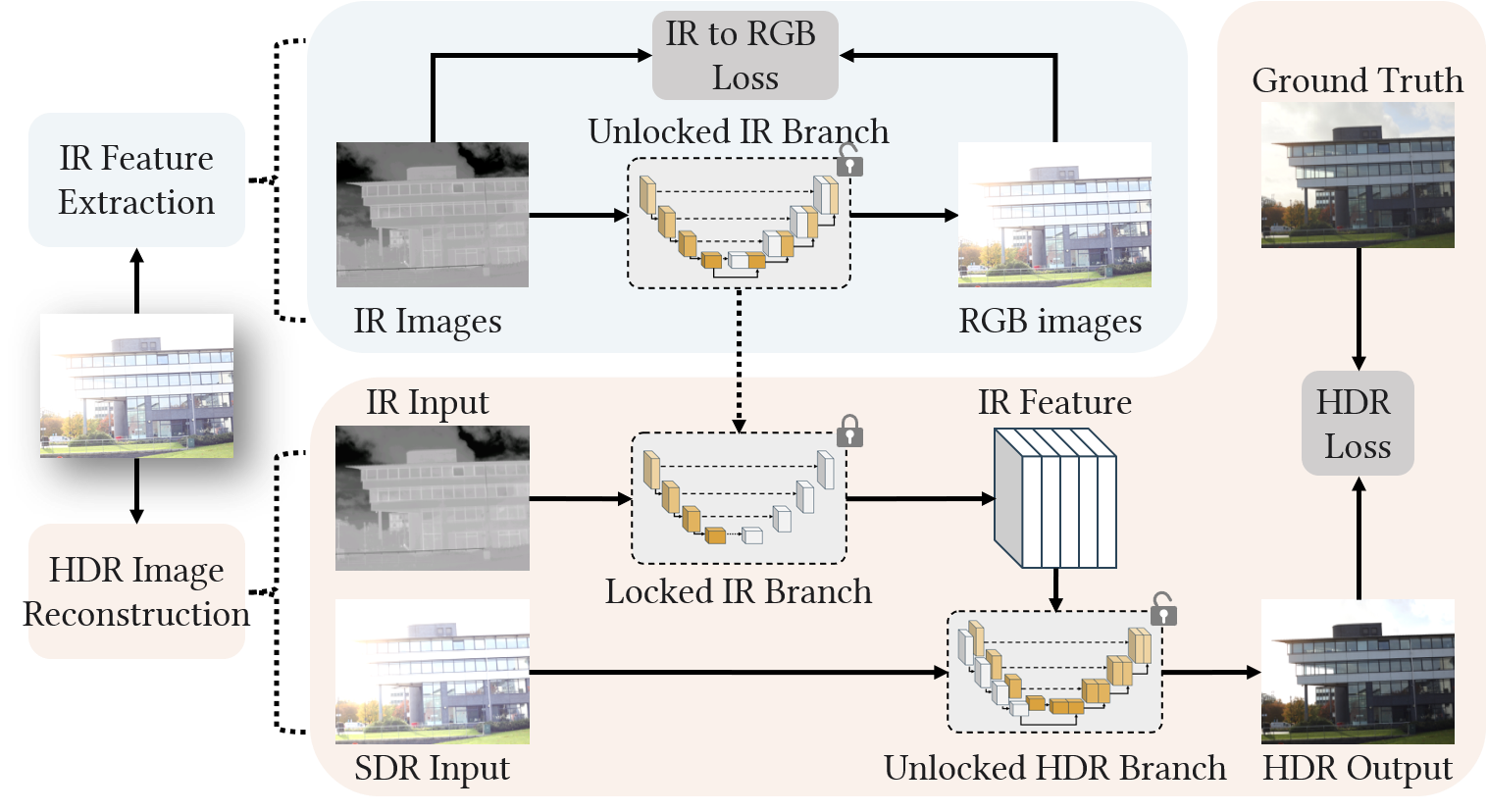} 
\caption{The full pipeline of the proposed HDRTNet for enhancing SDR images using thermal to obtain HDR images.}
\label{fig:The-overall-process}
\vspace{-0.3cm}
\end{figure}

The proposed method consists of two steps to combine visible and IR information for generating HDR images: 1) infrared feature extraction and 2) HDR image reconstruction; see Fig. \ref{fig:The-overall-process}. Infrared feature extraction aims to extract information from infrared images to fill in missing information in SDR images. However, due to the difference between infrared and visible light modalities, the information from IR images is not always useful for HDR synthesis tasks. A simple example is that even if objects with different materials but the same color (such as a wooden white wall and an iron white wall) look similar in RGB images, they have significant differences in infrared images as their temperatures may be different. Therefore, to extract the features relevant to the visual task, we first train the IR branch to generate RGB images from IR images and use this for extraction of \emph{relevant} infrared features.

Once this is trained, the IR branch is frozen for the HDR imaging process, and is used to provide the features from the IR domain for use in HDR reconstruction. This is achieved by combining the IR features and SDR image in the HDR branch to synthesize an HDR image.

One might ask why we use this two step approach rather than directly combining RGB and IR information at the pixel level. We initially explored this approach, and as discussed in the ablation study, this leads to registration errors from the use of multiple sensors; something that the proposed two step approach avoids.\\

\begin{figure}[!t]
\centering
\includegraphics[width=0.7\textwidth]{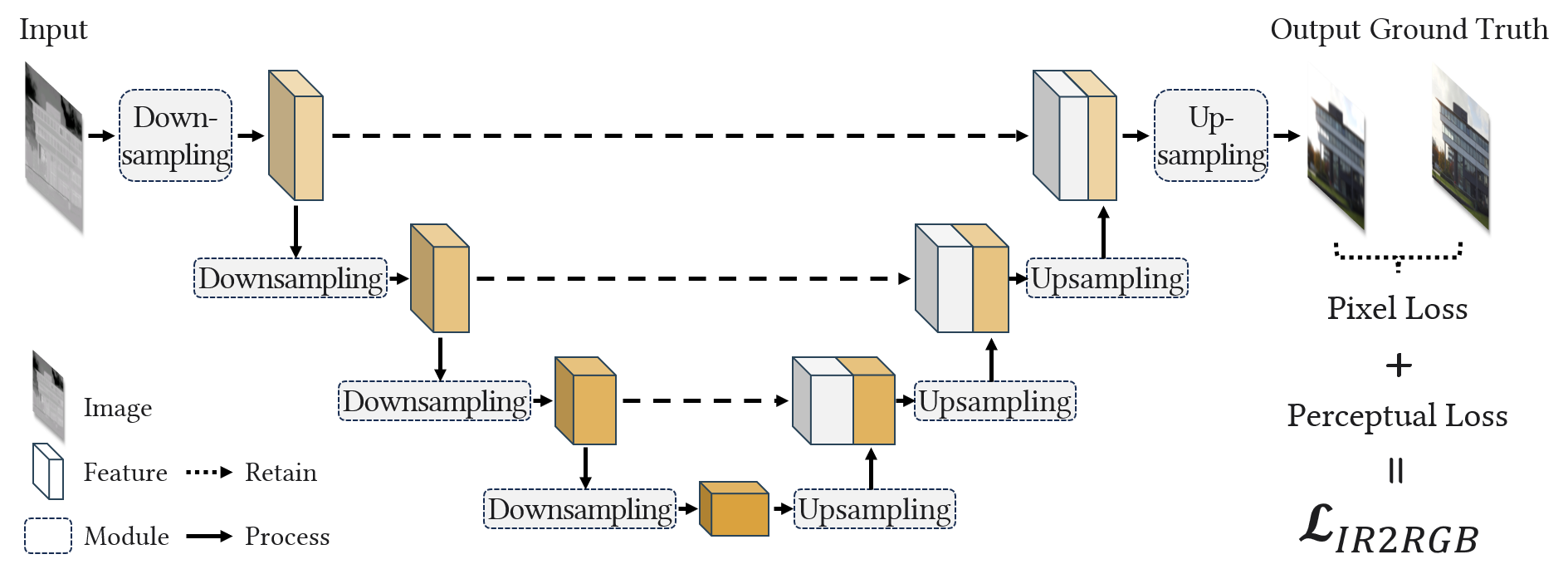} 
\caption{\label{fig:The-structure-of-IR}The structure of the IR branch.}
\end{figure}

\noindent\textbf{Infrared Feature Extration} In the infrared feature extraction process, the IR branch learns to
convert IR images to RGB images. The structure of the IR branch can be seen in Fig. \ref{fig:The-structure-of-IR}, and employs a U-Net architecture. This structure was chosen as we require a mechanism to extract relevant features from the IR data while mapping single channel IR data to three channel RGB data; a U-Net architecture satisfies this goal.

This U-Net structure consists of four downsampling and four upsampling modules.
Each downsampling module has a max pooling layer and a double convolution layer. Inside the double convolution layer, there are two instances of the same combination structure: a convolution layer, a Batch Normalization layer, and a rectified linear unit (ReLU) layer.
Each upsampling module has a transposed convolution layer and a double convolution layer.

The loss function of the IR branch is defined as: 
\begin{equation}
\mathcal{L}_{IR2RGB}=\mathcal{L}_{pix}+\alpha\mathcal{L}_{per},
\end{equation}
\noindent where $\mathcal{L}_{pix}$ is the pixel loss and $\mathcal{L}_{per}$ is the perceptual loss. The pixel loss is used to minimize the difference between generated and real pixels. $\mathcal{L}_{pix}$ combines the mean absolute error (MAE, L1 loss) and cosine similarity. This is designed to minimize both intensity and color differences:
\begin{equation}
\mathcal{L}_{pix}=0.99\times\left\Vert GT-Out\right\Vert+0.01\times\frac{GT\cdot Out}{\left\Vert GT\right\Vert\times\left\Vert Out\right\Vert},
\end{equation}
\noindent where $Out$ is the generated HDR image and  $GT$ is the ground truth. The weights chosen are similar to other HDR reconstruction methods such as \cite{ExpandNet}. 
The perceptual loss uses the well-trained networks (VGG-19 in our work) to compare the L1 distance of output and ground truth in the feature space to improve the similarity between the generated RGB images and the real RGB images.
The weight $\alpha=1.0$, which is the same as the HDR branch. \\

\noindent\textbf{HDR Image Reconstruction \label{subsec:HDR-Image}}

\begin{figure}[!t]
\noindent \begin{centering}
\includegraphics[width=0.7\textwidth]{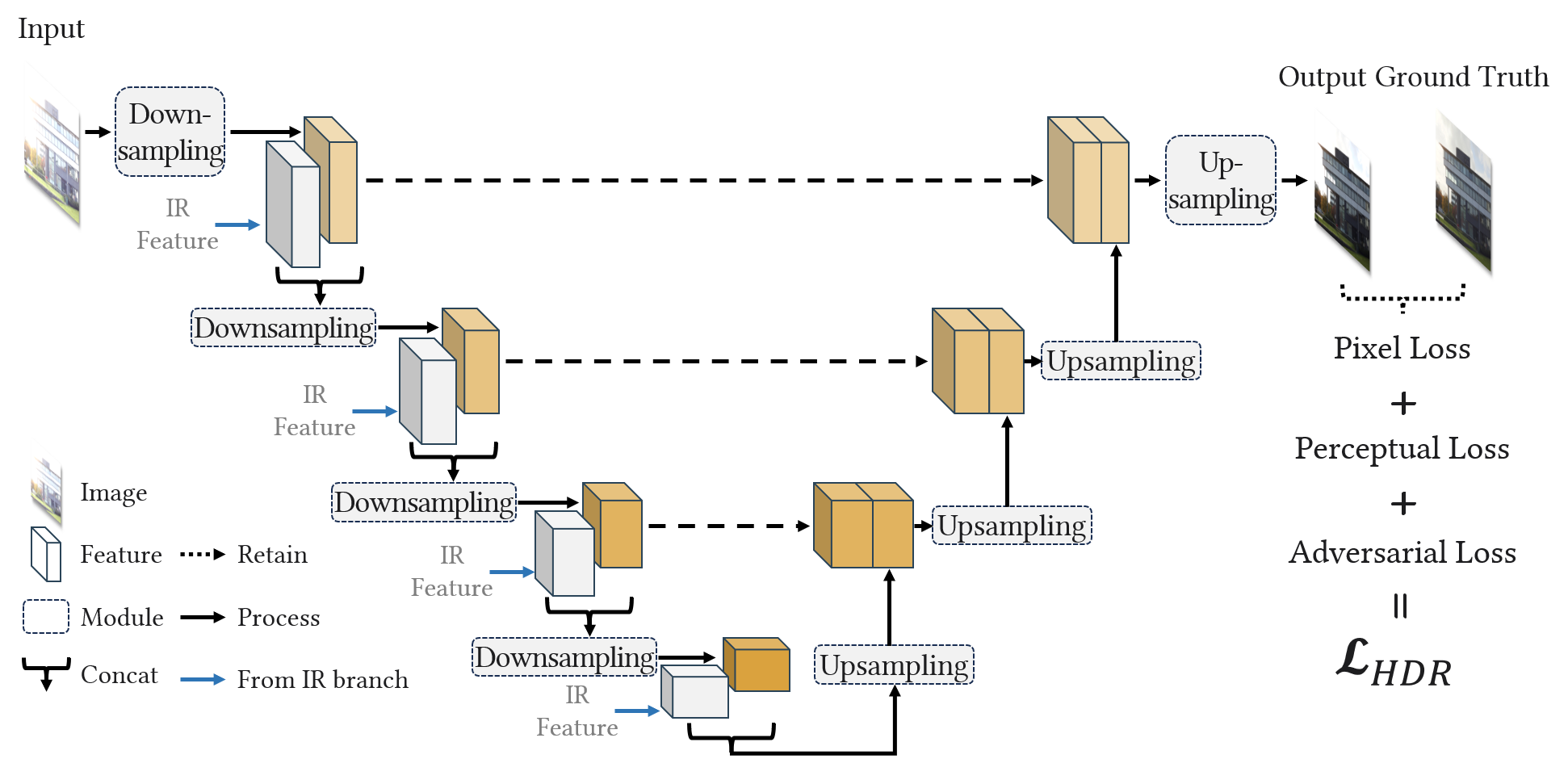} 
\par\end{centering}
\caption{\label{fig:The-structure-of-HDR}The structure of the HDR branch.}
\end{figure}

In the HDR image generation process, the HDR branch fuses the features extracted from the IR branch and generates HDR images from SDR images. The structure of HDR branch can be seen in Fig. \ref{fig:The-structure-of-HDR}. For similar reasons to the IR branch, the HDR branch utilizes a U-Net-like structure. The main difference is that before each downsampling module, the HDR branch concatenates the features from the former layer of the HDR branch and the features extracted from the IR branch. This aims to fuse the IR and visible information to extract features from both which are relevant to the HDR reconstruction task. During upsampling, only the features from the HDR branch will be retained and fed into the upsampling modules. This means that fusion is only implemented in the shallow layers, which can minimize the effects of differences between IR and RGB modalities.

In the training phase, the IR branch is first trained to learn to reconstruct RGB images. After that, the input layer and the first three downsampling modules of the IR branch are frozen and used in the HDR branch to extract IR features.
In the inference phase, the HDR branch (including the input layer and the first three downsampling modules of the IR branch) is used to reconstruct HDR images from SDR and IR images.

The loss function used for training the HDR branch, includes the pixel loss and perceptual loss introduced in the IR branch, but also also adopts an adversarial loss $\mathcal{L}_{GAN}$ \cite{GAN}. The adversarial loss allows the HDR branch to learn the distribution of real HDR images,  which aims to reduce artifacts and distortion in the generated images.
The overall loss function of the HDR branch can be expressed as:
\begin{equation}
\mathcal{L}_{HDR}=\mathcal{L}_{pix}+\alpha\mathcal{L}_{per}+\beta\mathcal{L}_{GAN},
\end{equation}
\noindent where the weights $\alpha$ and $\beta$ are determined experimentally (please refer to the supplementary for more details). In our work, $\alpha=1.0$ and $\beta=1\textrm{e}-5$.
More details about the experiments related to weight determination are presented in the \textit{Ablation Study}.

\begin{table*}[!t]
\caption{\label{tab:The-results-of}The results of the HDRT dataset. The best
results are highlighted in \textcolor{red}{red}, and the second-best
results in \textcolor{orange}{orange}. The table is divided into
three subgroups: 1) overexposed images, 2) underexposed images,
3) all images.}
\centering{}\resizebox{\textwidth}{!}{
\begin{tabular}{lccc|ccc|ccc}
\toprule
\multirow{2}{*}{Methods} & \multicolumn{3}{c|}{Over Exposure} & \multicolumn{3}{c|}{Under Exposure} & \multicolumn{3}{c}{All}\tabularnewline
\cmidrule(l){2-10} & pu-PSNR & pu-SSIM & pu-VSI & pu-PSNR & pu-SSIM & pu-VSI & pu-PSNR & pu-SSIM & pu-VSI
\tabularnewline
\midrule
DrTMO \cite{End} & 30.515 & 0.544 & 0.959 & 36.574 & 0.849 & 0.983 & 33.545 & 0.696 & 0.971
\tabularnewline
Deep Recursive HDRI \cite{DeepRecursiveHDRI} & 32.997 & 0.728 & 0.959 & 41.139 & 0.938 & 0.988 & 37.068 & 0.833 & 0.974
\tabularnewline
HDRTVNet \cite{HDRTVNet} & 31.074 & 0.633 & 0.953 & 39.739 & 0.802 & 0.984 & 35.407 & 0.717 & 0.969
\tabularnewline
LaNet \cite{LaNet} & 33.580 & 0.681 & 0.965 & 40.120 & 0.919 & 0.985 & 36.850 & 0.800 & 0.975
\tabularnewline
HDRCNN \cite{Eil} & 35.515 & 0.703 & \textcolor{orange}{0.966} & 42.737 & 0.872 & 0.988 & 39.126 & 0.788 & 0.977
\tabularnewline
\multicolumn{1}{l}{Deep-HDR Reconstruction \cite{SAN}} & 30.909 & 0.554 & 0.955 & 43.001 & 0.880 & 0.990 & 36.955 & 0.717 & 0.972
\tabularnewline
ICTCPNet \cite{ICTCPNet} & 21.371 & 0.253 & 0.938 & 31.271 & 0.629 & 0.974 & 26.321 & 0.441 & 0.956
\tabularnewline
ExpandNet \cite{ExpandNet} & 25.633 & 0.415 & 0.938 & 39.288 & 0.882 & 0.985 & 32.461 & 0.649 & 0.962
\tabularnewline
HDRUNet \cite{HDRUNet} & \textcolor{orange}{39.216} & \textcolor{orange}{0.893} & 0.964 & \textcolor{orange}{44.270} & \textcolor{orange}{0.947} & \textcolor{orange}{0.991} & \textcolor{orange}{41.743} & \textcolor{orange}{0.920} & \textcolor{orange}{0.978}
\tabularnewline
HDRTNet (Ours) & \textcolor{red}{40.589} & \textcolor{red}{0.922} & \textcolor{red}{0.977} & \textcolor{red}{45.359} & \textcolor{red}{0.949} & \textcolor{red}{0.994} & \textcolor{red}{42.974} & \textcolor{red}{0.936} & \textcolor{red}{0.985}
\tabularnewline
\bottomrule
\end{tabular}}
\end{table*}

\subsection{Comparison with Other Methods}

We trained and tested the proposed method using PyTorch on an
Intel Xeon W-2295 CPU and Nvidia Quadro 8000 GPU. We used Adam, with a learning rate of 4e-5, and 200 epochs. We used the MultiStepLR scheduler, and the learning rate was reduced by half every 20000 steps. The training set comprises of 8000 images; the validation set consists of 2000 images. To evaluate the performance compared to other methods, we utilized the perceptually uniform (pu) version of three commonly used metrics: peak signal-to-noise ratio (pu-PSNR), structural similarity index measure (pu-SSIM), and the visual saliency index (pu-VSI). These perceptually uniform metrics \cite{PU21} are used as linear high dynamic range color values are non-linearly related to our perception of visible differences, so we converted absolute high dynamic range linear color values into perceptually uniform values.

As our approach is designed for single time capture (i.e. visible and IR information is captured simultaneously), we compared with the state-of-the-art single image HDR reconstruction methods and viewed multi frame capture approaches as being out of scope (e.g. multiple exposure video reconstruction approaches or flash/no flash pairs). We therefore compared with the following methods: DrTMO \cite{End}, Deep Recursive HDRI \cite{DeepRecursiveHDRI}, HDRTVNet \cite{HDRTVNet}, LaNet \cite{LaNet}, HDRCNN \cite{Eil}, Deep-HDR Reconstruction \cite{SAN}, ICTCPNet \cite{ICTCPNet}, ExpandNet \cite{ExpandNet}, and HDRUNet \cite{HDRUNet}. Tab. \ref{tab:The-results-of} presents the results of our HDRT dataset. Our analysis divided the table into three groups of images where HDR imaging is required, over-exposed images, under-exposed images, and a combination of these images. The proposed method obtains the best performance in all three groups.

\begin{figure}[!ht]
\centering
\includegraphics[width=0.8\textwidth]{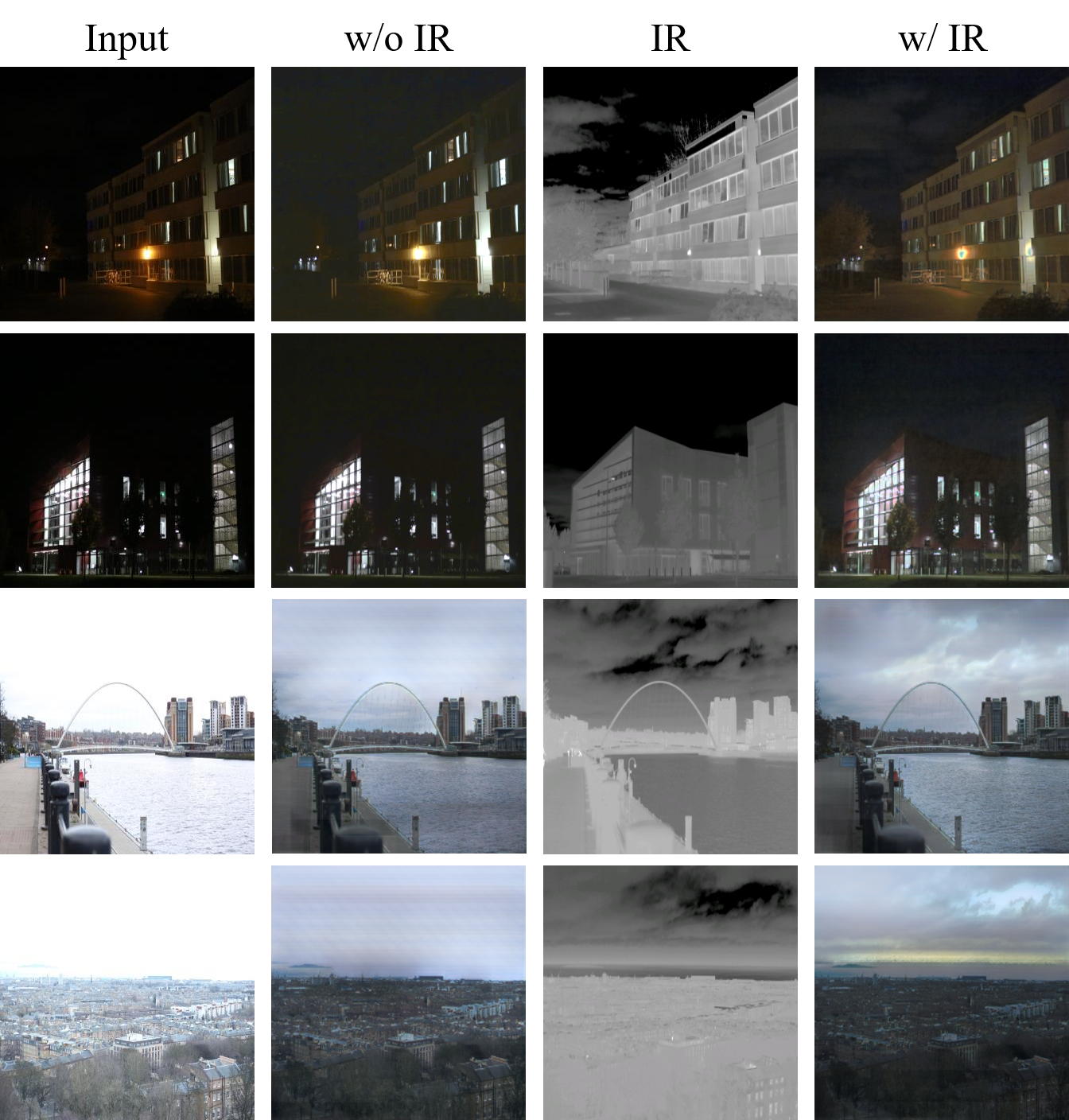}
\caption{\label{fig:Examples-of-our}Examples of HDRTNet with and without infrared images.}
\end{figure}

\begin{table*}[!t]
\caption{\label{tab:Ablation-study-of}Ablation study of the proposed method on our HDRT dataset. ``\textsurd'' indicates the technique is adopted, ``\texttimes'' indicates not, and ``-'' means not applicable.}
\centering{}\resizebox{\textwidth}{!}{
\begin{tabular}{lcccccc|ccc|ccc}
\toprule
\multirow{2}{*}{Methods} & \multirow{2}{*}{IR images} & Feature-level & IR Branch & \multicolumn{3}{c|}{Over Exposure} & \multicolumn{3}{c|}{Under Exposure} & \multicolumn{3}{c}{All
}\tabularnewline
\cmidrule(l){5-13} 
 &  & fusion & separately training & pu-PSNR & pu-SSIM & pu-VSI & pu-PSNR & pu-SSIM & pu-VSI & pu-PSNR & pu-SSIM & pu-VSI
\tabularnewline
\midrule
RGB & \texttimes{} & - & - & 39.445 & 0.899 & 0.962 & 44.614 & 0.940 & 0.993 & 42.029 & 0.920 & 0.978
\tabularnewline
Pixel & \textsurd{} & \texttimes{} & \texttimes{} & 40.156 & 0.914 & 0.972 & 44.732 & 0.947 & 0.993 & 42.444 & 0.930 & 0.983
\tabularnewline
Combined (Naive) & \textsurd{} & \textsurd{} & \texttimes{} & 40.194 & 0.914 & 0.972 & 44.962 & 0.948 & 0.993 & 42.578 & 0.931 & 0.983
\tabularnewline
HDRTNet (Ours) & \textsurd{} & \textsurd{} & \textsurd{} & \textbf{40.589} & \textbf{0.922} & \textbf{0.977} & \textbf{45.359} & \textbf{0.949} & \textbf{0.994} & \textbf{42.974} & \textbf{0.936} & \textbf{0.985}
\tabularnewline
\bottomrule
\end{tabular}}\vspace{-0.3cm}
\end{table*}

In the overexposure case, the proposed HDRTNet outperforms the second-best HDRUNet by 1.373 and 0.029 in pu-PSNR and pu-SSIM, respectively. Additionally, HDRTNet outperformed the second-best HDRCNN by 0.011 in pu-VSI. In the underexposure case, HDRTNet outperformed the second-best HDRUNet by 1.089, 0.002, and 0.003 in pu-PSNR, pu-SSIM, and pu-VSI, respectively. Overall, these results show that HDRTNet outperformed the second-best method by 1.231, 0.016, and 0.007 on the pu-PSNR, pu-SSIM, and pu-VSI, respectively.

\subsection{Ablation Study\label{subsec:Ablation-Study} }

To demonstrate the effectiveness of our major modules, we implemented three variations: baseline RGB without IR, pixel-level rather than feature-level fusion, and combined rather than separate training of the branches. Baseline RGB without IR refers to only using a standard U-Net \cite{UNet} to recover HDR images from SDR images (referred to as ``RGB''). Pixel-level fusion also uses a standard U-Net, but different from RGB this method adopts IR image data as the fourth channel of the input, from which the network can extract information from the IR domain (referred to as ``Pixel''). Combined training refers to fusing the features from IR and SDR images in feature space, but training together with the HDR branch rather than separately training and reconstructing RGB images (referred to as ``Combined (Naive)'').

The comparison results on the HDRT dataset are shown in Tab. \ref{tab:Ablation-study-of}. 
Since the feature-level fusion and IR branch cannot operate with only RGB, we excluded these methods.\\

\noindent\textbf{RGB Baseline.} Comparing ``Pixel'' with the baseline ``RGB'', Pixel exceeds RGB in pu-PSNR, pu-SSIM, and pu-VSI by 0.415, 0.011, and 0.005, respectively.
This is because, as discussed before, due to sensor limitations SDR images lose information compared with HDR images.
The inclusion of data from the IR domain adds more details, as is illustrated in Fig. \ref{fig:Examples-of-our}. \\

\begin{figure}[!t]
\begin{centering}
\subfloat[Visual comparison between RGB, Pixel, and our HDRTNet.]{\centering\includegraphics[width=0.8\textwidth]{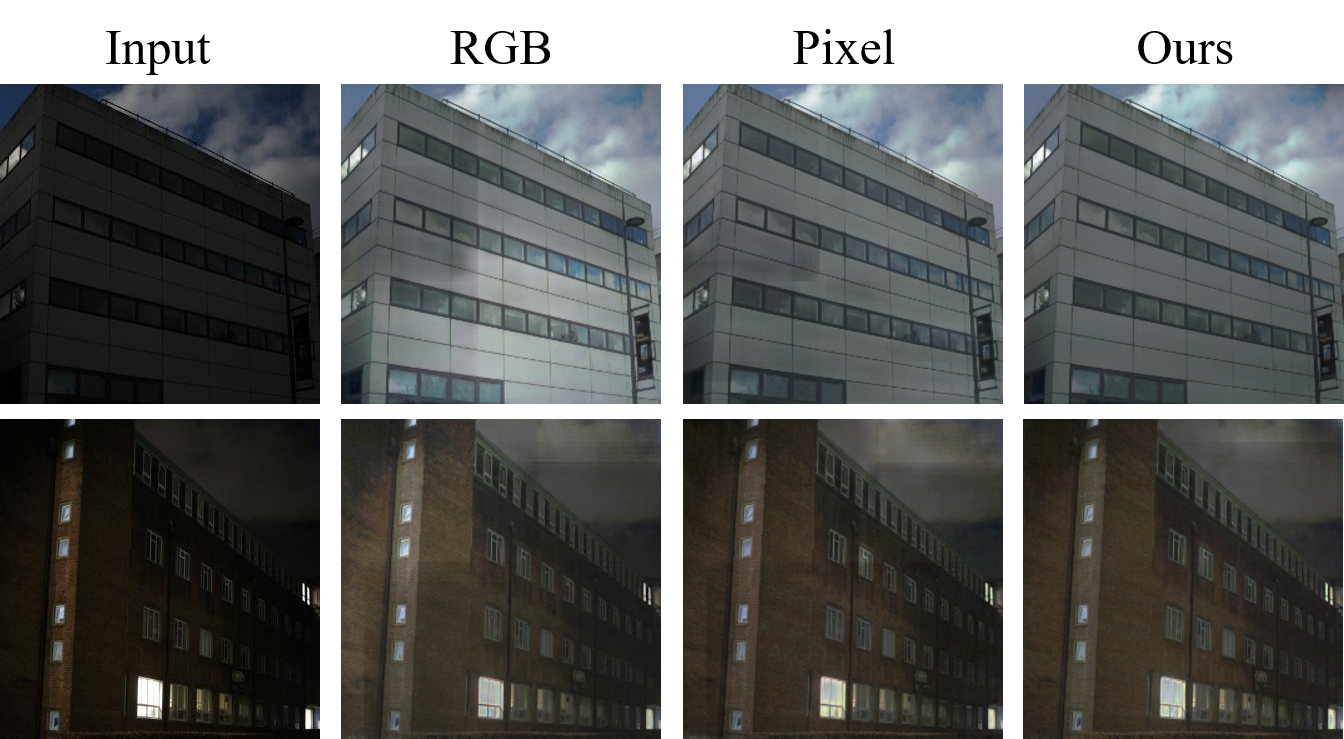}}
\par
\end{centering}\begin{centering}
\subfloat[Examples of ghosting caused by registration errors.]{\centering\includegraphics[width=0.8\textwidth]{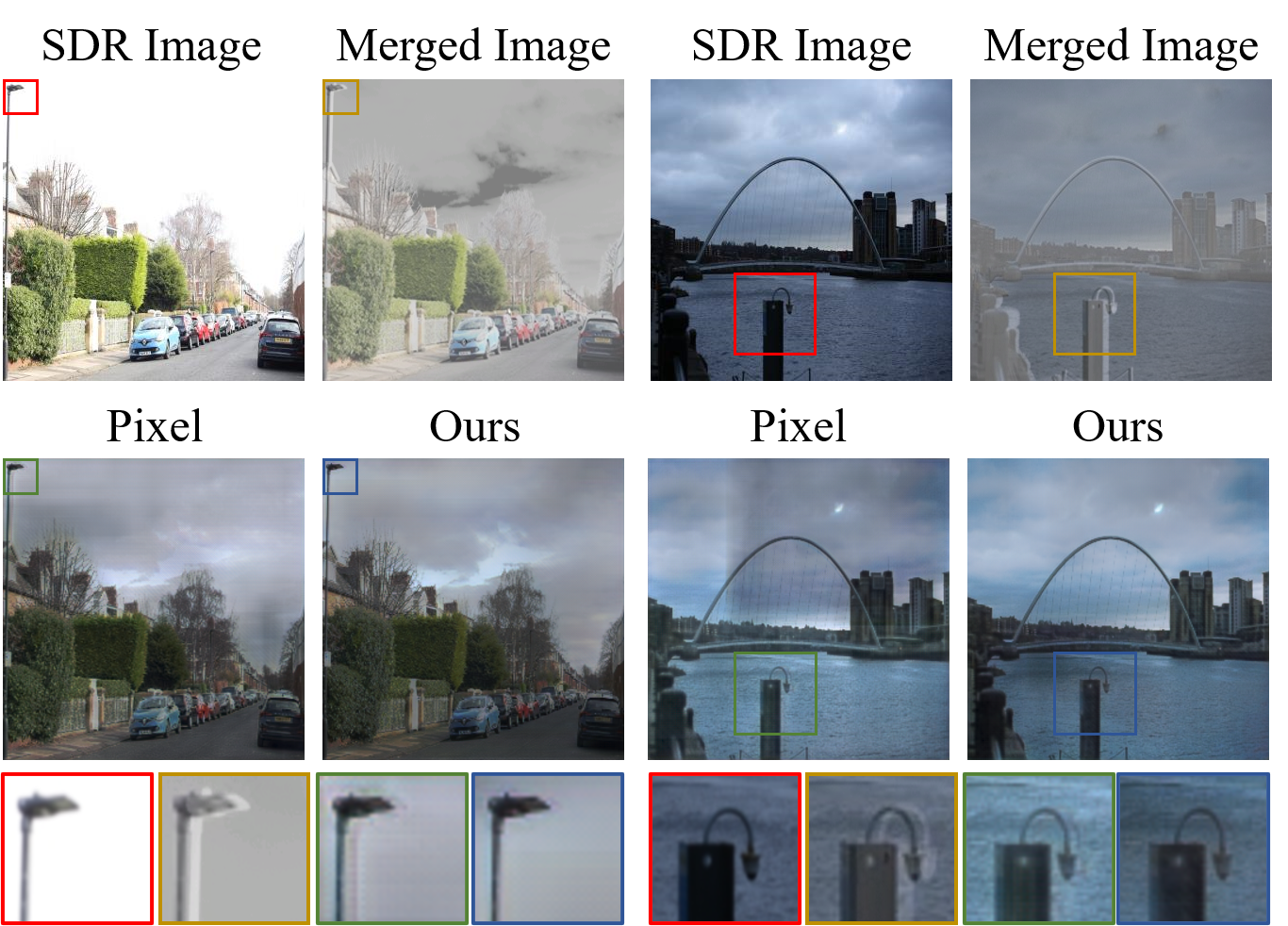}}
\par\end{centering}
\caption{\label{fig:Visual-comparison-between}Visual comparison between the pixel-level fusion and our HDRTNet.}
\end{figure}

\noindent\textbf{Pixel-Level Fusion.} However, comparing HDRTNet with ``Pixel'', HDRTNet still exceeds Pixel in pu-PSNR, pu-SSIM, and pu-VSI by 0.530, 0.005, and 0.003, respectively.
This is because, without a separate sub-network handling IR image data, the pixel-level fusion is prone to ignoring heterogeneous regions in IR images, resulting in uncompleted info from IR domain being extracted. 
As shown in Fig. \ref{fig:Visual-comparison-between} (a), Pixel is visually better than RGB but still has some artifacts compared to HDRTNet.

Another problem with pixel-level fusion is that it is not robust enough for binocular camera registration errors. As shown in Fig. \ref{fig:Visual-comparison-between} (b), IR and RGB images are not completely registered, and a clear deviation in their positions can be seen in the second column of fused images.
The results of Pixel clearly show this ghosting, but HDRTNet does not. As a result, pixel-level fusion produces more blurry images due to tiny registration errors.\\

\begin{figure}[!t]
\begin{centering}
\subfloat[Visual comparison between Pixel, Combined, and HDRTNet.]{\includegraphics[width=0.8\textwidth]{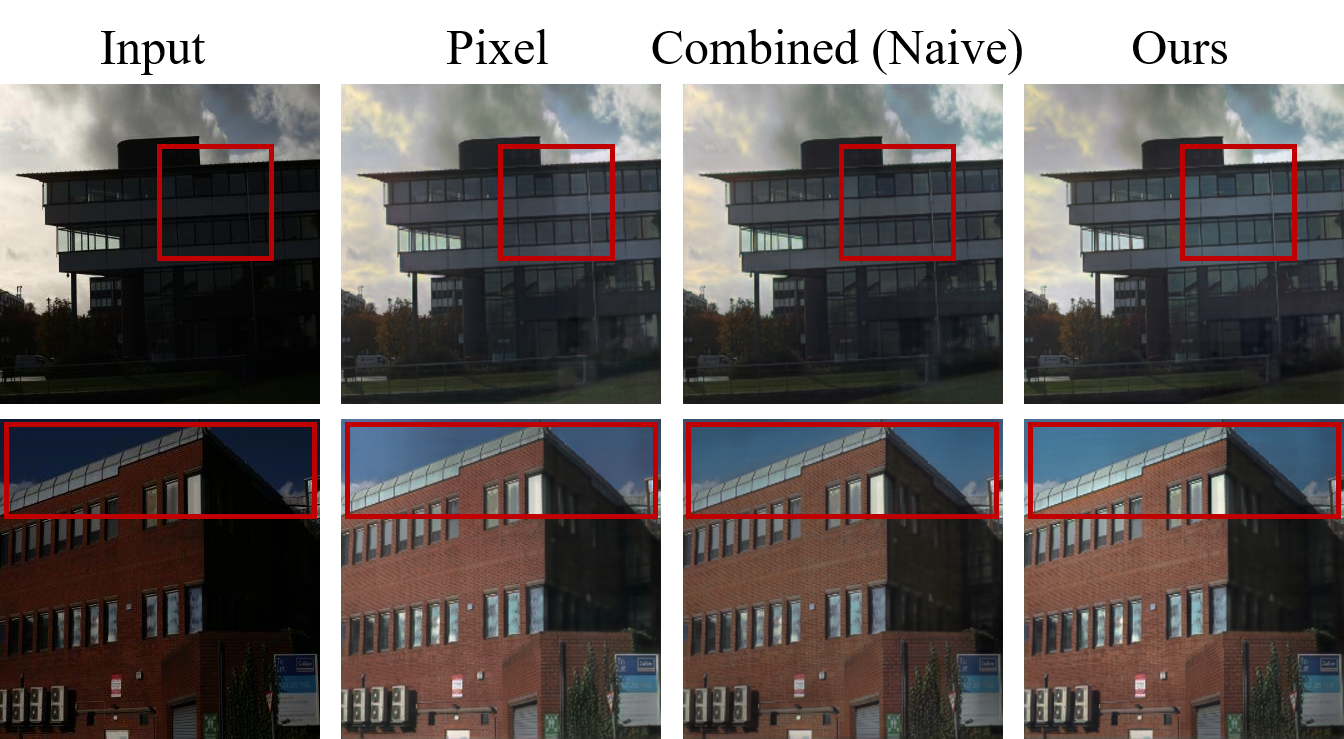}}
\par\end{centering}
\begin{centering}
\subfloat[Examples of color shift caused by visual-irrelevant features.]{\includegraphics[width=0.8\textwidth]{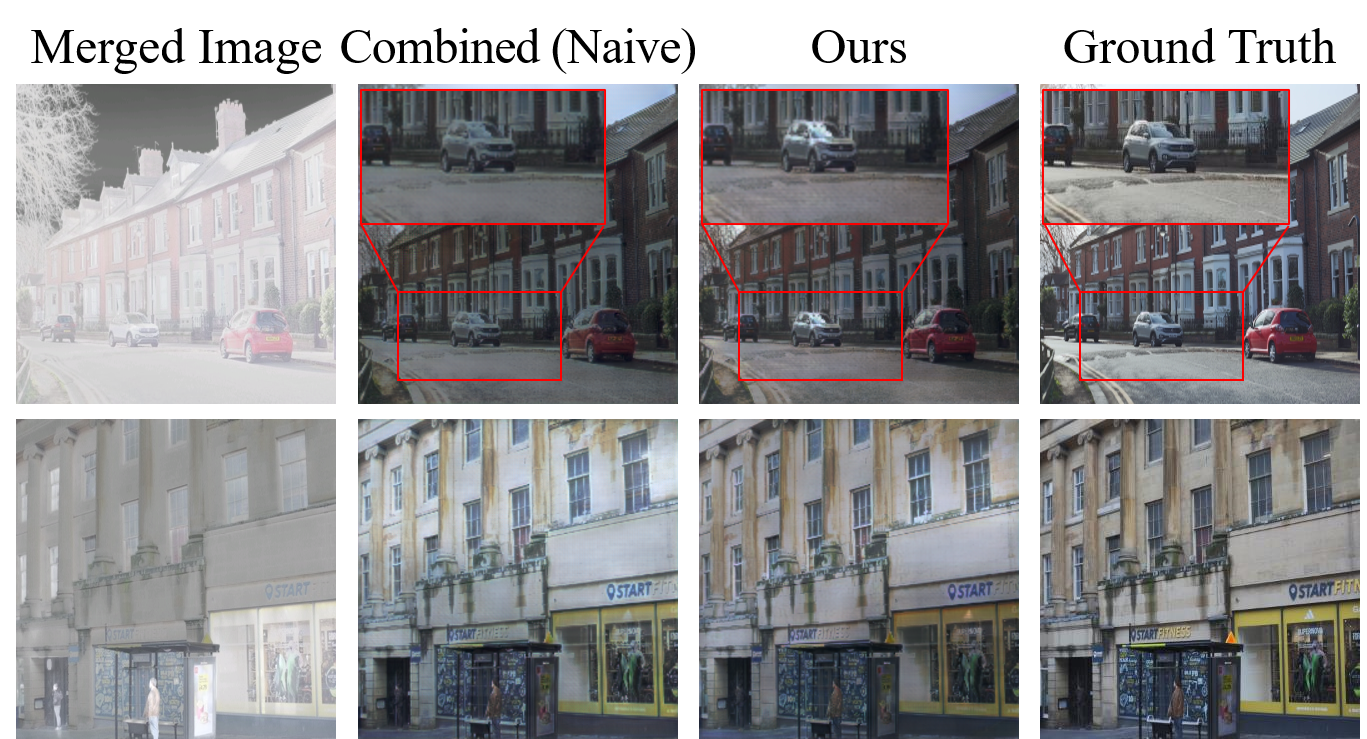}}
\par\end{centering}
\caption{\label{fig:Visual-comparison-between-feature}Visual comparison between the feature-level fusion and HDRTNet.}
\vspace{-0.3cm}
\end{figure}

\noindent\textbf{Naive Combined Training.} Comparing ``Combined (Naive)'' with ``Pixel'', Combined exceeds Pixel in pu-PSNR and pu-SSIM by 0.134 and 0.001, respectively, which means that the feature-level fusion can help resolve the problem of output instability. 
As seen in Fig. \ref{fig:Visual-comparison-between-feature} (a), the output of Combined is very smooth and does not produce artifacts like Pixel. 
However, HDRTNet still exceeds Combined in pu-PSNR, pu-SSIM, and pu-VSI by 0.396, 0.005, and 0.002, respectively. 
This is because visually irrelevant features reduce the resulting quality of the HDR image, as shown in Fig. \ref{fig:Visual-comparison-between-feature} (b).
In Fig. \ref{fig:Visual-comparison-between-feature} (b), the output from Combined is less saturated compared to the ground truth, while HDRTNet more closely resembles the color information present.
Compared with the ground truth, the image generated by Combined is more like a fusion of the RGB and IR images (the first column of Fig. \ref{fig:Visual-comparison-between-feature} (b)), in which the colors are whiter and simpler than ground truth images.
However, the IR branch in HDRTNet is trained to reconstruct RGB images, which can avoid extracting visual-irrelevant features. Therefore, HDRTNet is not easily affected by visually irrelevant features in the IR domain.\\

\noindent\textbf{Loss Weight.} In order to test the impact of different weights $\alpha$ and $\beta$ on HDR performance, we use HDRTNet with different weights to train and reconstruct HDR images; the results are shown in Tab. \ref{tab:Ablation-alpha-beta}.
The results show that the performance is best when $\alpha$ is set to 1.0 and $\beta$ is set to $1\textrm{e}-5$. 

\begin{table}[!t]
\caption{\label{tab:Ablation-alpha-beta} Ablation study on $\alpha$ and $\beta$.}
\centering{}
\begin{tabular}{ccccc}
\toprule
$\alpha$ & $\beta$ & pu-PSNR & pu-SSIM & pu-VSI\tabularnewline
\midrule 
e-1 & e-6 & 44.292 & 0.933 & 0.991\tabularnewline
e-1 & e-5 & 44.516 & 0.942 & 0.991\tabularnewline
e-1 & e-4 & 44.292 & 0.933 & 0.991\tabularnewline
1 & e-6 & 44.802 & 0.940 & 0.991\tabularnewline
1 & e-5 & \textbf{45.359} & \textbf{0.949} & \textbf{0.994}\tabularnewline
1 & e-4 & 44.811 & 0.940 & 0.991\tabularnewline
e+1 & e-6 & 44.506 & 0.936 & 0.991\tabularnewline
e+1 & e-5 & 44.435 & 0.934 & 0.991\tabularnewline
e+1 & e-4 & 44.895 & 0.943 & 0.991\tabularnewline
\bottomrule 
\end{tabular}
\end{table}

\subsection{Visual Performance}

\begin{figure*}[!p]
\hspace{-0.5cm}
\includegraphics[width=1.0\textwidth]{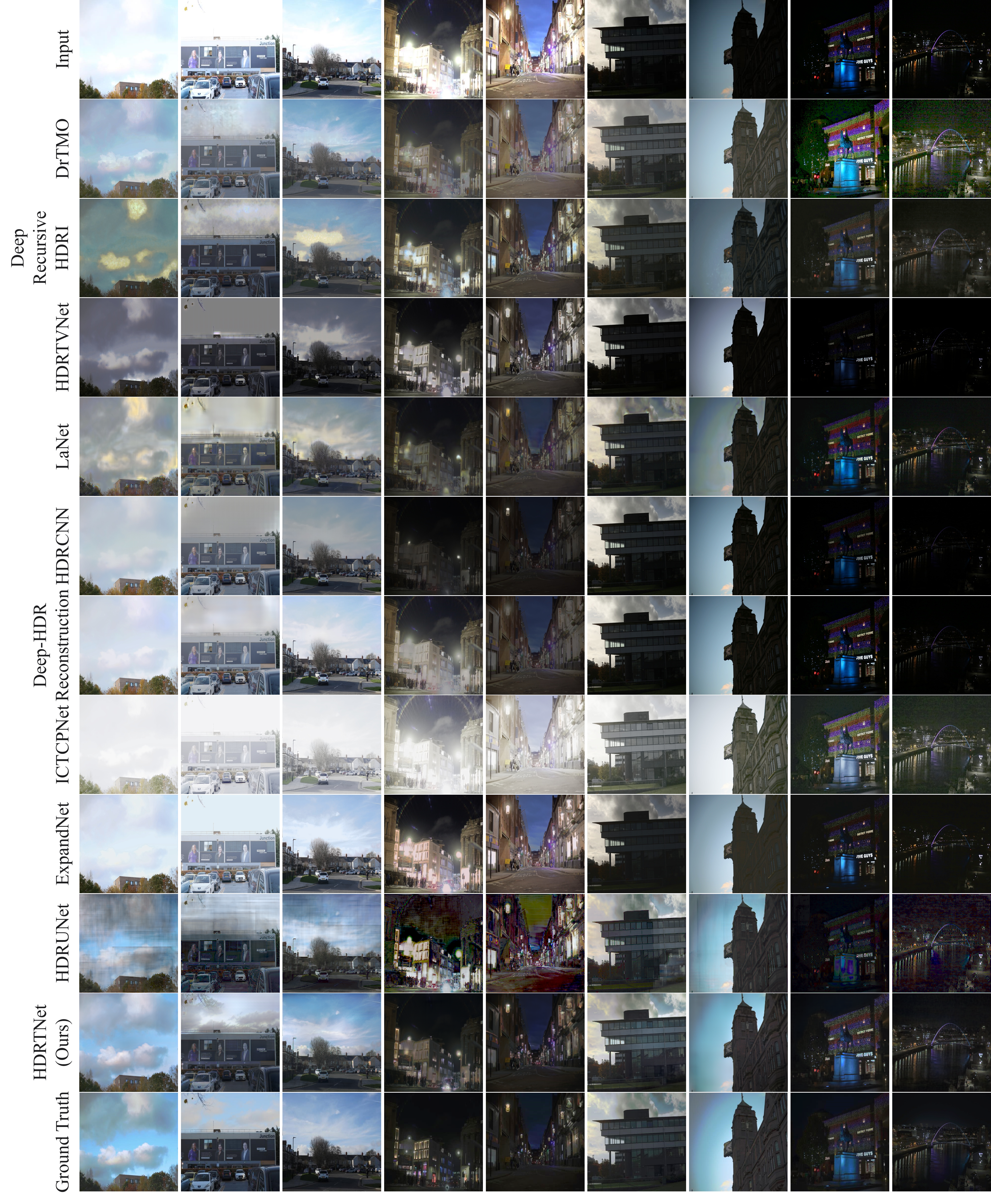}
\vspace{-0.3cm}
\caption{\label{fig:A-visual-comparison}A visual comparison of all tested
methods (tone-mapped).}
\end{figure*}

This section demonstrates qualitative results by comparing the proposed HDRTNet with the state-of-the-art and the original HDR frames.
The input for all methods are the SDR images.
All results are grouped by images containing mostly over-exposed and under-exposed regions. 
The results can be seen in Fig. \ref{fig:A-visual-comparison}.
All HDR images use the Durand tone mapping operator \cite{Durand} for a fair visual comparison.

The first five columns in Fig. \ref{fig:A-visual-comparison} show
the over-exposure cases. The main challenge in these scenarios is
to recover cloud details in the first three columns and to restore missing details around the lights in the fourth and fifth columns. With additional information from the IR data, HDRTNet produces images closest to the ground truth image. 

In Fig. \ref{fig:A-visual-comparison}, the last four columns show
under-exposure cases, which makes it challenging to restore the building details in the sixth and seventh columns and the dark details of the nighttime images in the eighth and last columns. 
Again, HDRTNet utilizes IR information to generate HDR images closes to the ground truth images. 

\begin{figure*}[!p]
\hspace{-0.5cm}
\includegraphics[width=0.96\textwidth]{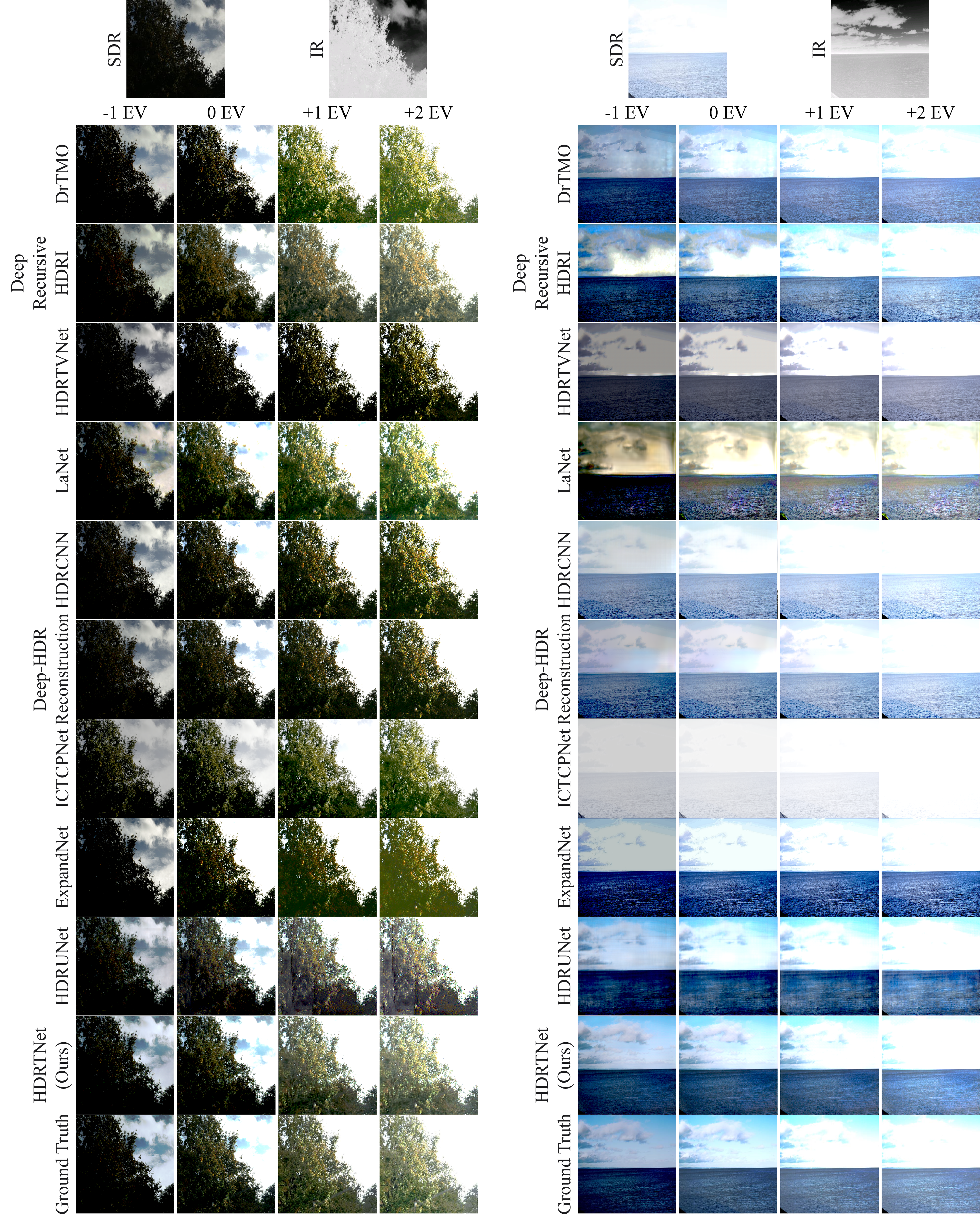}
\vspace{-0.3cm}
\caption{\label{fig:Another-visual-comparison1}A visual comparison of all tested methods (exposure blankets).}
\end{figure*}

We also present a series of further results showcasing HDR images under various exposure settings.
Included are comparisons between HDR images recovered by the proposed HDRTNet, the ground-truth HDR images, and counterparts generated by other state-of-the-art ITMOs.
The results can be seen in Fig. \ref{fig:Another-visual-comparison1}. Our approach is closest to ground truth across all exposure values.

\begin{figure*}[!p]
\includegraphics[width=1.0\textwidth]{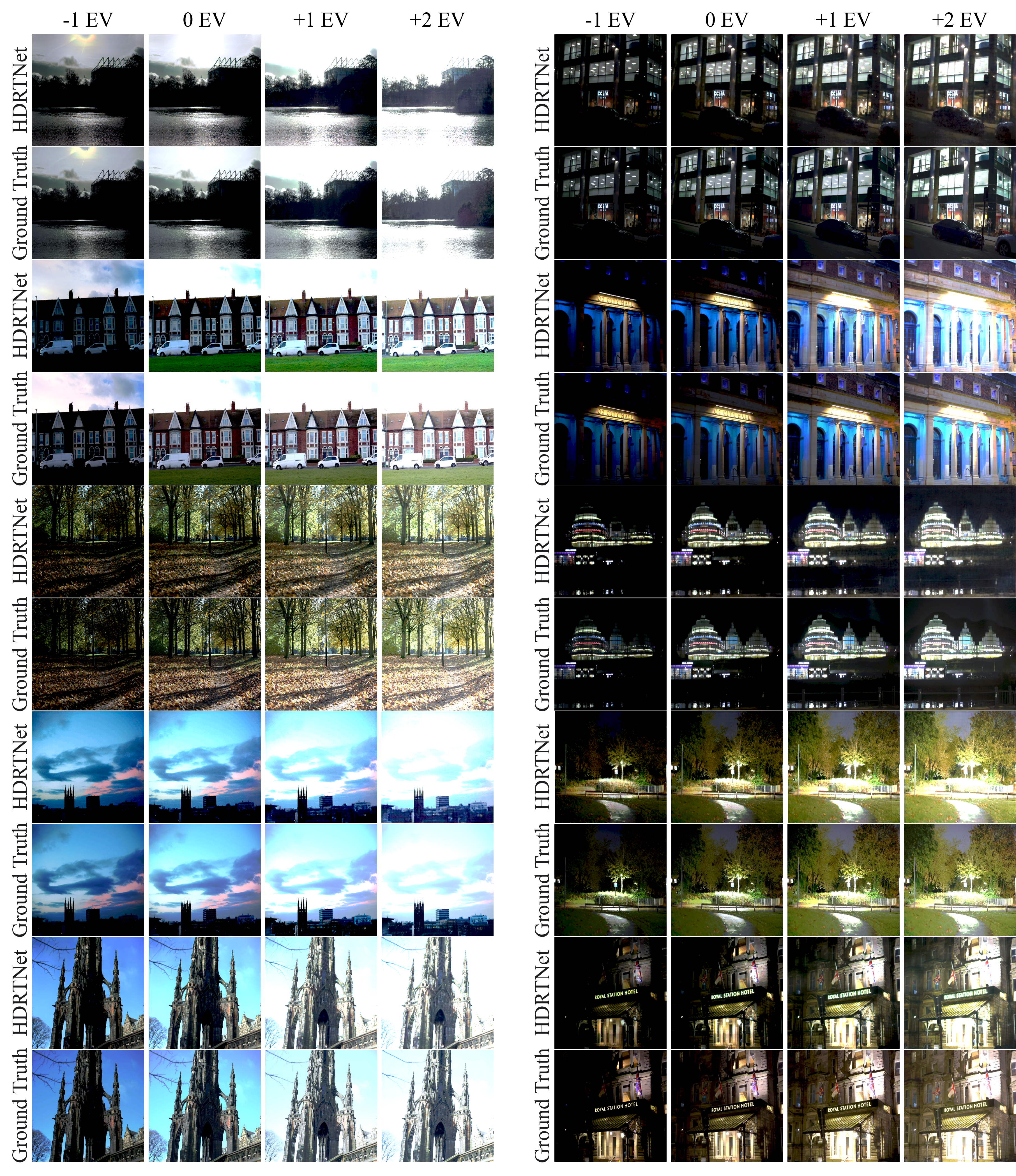}
\caption{\label{fig:Another-visual-comparison2}Visual results of the proposed HDRTNet in different scenarios.}
\end{figure*}

Furthermore, the effectiveness of the proposed HDRTNet can be extended to a diverse array of scenes, highlighting the robust generalization capability across different scenarios, as is shownin Fig. \ref{fig:Another-visual-comparison2}.
These visual results demonstrate the strong HDR imaging performance of our HDRTNet and its broad applicability.

\subsection{Discussion}

Our proposed method performs well in enhancing HDR in over- or under-exposed conditions. However, we still need to consider how to utilize IR info to improve HDR when the lighting conditions are completely well-exposed. Furthermore, some materials block IR transmission while letting through visible light. Our method is likely to be less effective in these cases.
We currently use two cameras separately to capture IR and SDR images. Binocular errors may occur when there are far and near objects in the scenario, as shown in Fig. \ref{fig:Visual-comparison-between} (b). In such cases, precise registration of IR and RGB images is necessary. Alternatively, a new registration pre-processing model can be developed to ensure accurate registration before feeding two images into the network, or ultimately, the IR and RGB sensors can be integrated into one camera, which requires hardware improvements.

\section{Conclusion}
In this paper, we have introduced HDRT, a novel dataset for imaging which consists of 10K paired thermal infrared and visible light images.
Our dataset provides a comprehensive collection of images captured under diverse lighting conditions, which serves as a valuable resource for advancing HDR and IR imaging research.


Extensive experiments utilizing the HDRT dataset have demonstrated its effectiveness in improving HDR imaging techniques. 
Our proposed method which outperforms other state-of-the-art methods and, in particular, excels at well under- and over-exposed images, in which there is information loss due to adverse lighting environments and sensor limitations.
In addition, ablation experiments have demonstrated that our novel modules for feature-level fusion and infrared feature extraction can effectively extract infrared features and avoid HDR performance
degradation caused by modality differences between infrared and visible
light. 

The HDRT dataset not only provides a solid foundation for current IR-guided HDR imaging research but also holds significant potential for a wide range of applications, including other HDR methods, multi-modal fusion, domain adaptation, and advanced image reconstruction tasks.
To summarize, our work proposes and validates the use of thermal infrared imaging to capture HDR images, and we hope future work will build on our pipeline for HDR content capturing.


\clearpage

\bibliographystyle{elsarticle-num}
\bibliography{ref}

\begin{thebibliography}{10}
\expandafter\ifx\csname url\endcsname\relax
  \def\url#1{\texttt{#1}}\fi
\expandafter\ifx\csname urlprefix\endcsname\relax\def\urlprefix{URL }\fi
\expandafter\ifx\csname href\endcsname\relax
  \def\href#1#2{#2} \def\path#1{#1}\fi

\bibitem{HDR1}
Y.~Chen, G.~Jiang, M.~Yu, H.~Xu, Y.-S. Ho, \href{https://www.sciencedirect.com/science/article/pii/S1566253522001944}{Learning to simultaneously enhance field of view and dynamic range for light field imaging}, Information Fusion 91 (2023) 215--229.
\newblock \href {https://doi.org/https://doi.org/10.1016/j.inffus.2022.10.021} {\path{doi:https://doi.org/10.1016/j.inffus.2022.10.021}}.
\newline\urlprefix\url{https://www.sciencedirect.com/science/article/pii/S1566253522001944}

\bibitem{face_maching}
R.~Suma, K.~Debattista, D.~Watson, E.~Blagrove, A.~Chalmers, Subjective evaluation of high dynamic range imaging for face matching, IEEE Transactions on Emerging Topics in Computing 9~(4) (2021) 2042--2052.
\newblock \href {https://doi.org/10.1109/TETC.2019.2958738} {\path{doi:10.1109/TETC.2019.2958738}}.

\bibitem{fault_detection}
A.~R. Singh, T.~Bashford-Rogers, D.~Marnerides, K.~Debattista, S.~Hazra, Hdr image-based deep learning approach for automatic detection of split defects on sheet metal stamping parts, The International Journal of Advanced Manufacturing Technology 125~(5) (2023) 2393--2408.
\newblock \href {https://doi.org/10.1007/s00170-022-10763-6} {\path{doi:10.1007/s00170-022-10763-6}}.

\bibitem{HDR_detection}
E.~Onzon, F.~Mannan, F.~Heide, Neural auto-exposure for high-dynamic range object detection, in: Proceedings of the IEEE/CVF Conference on Computer Vision and Pattern Recognition (CVPR), 2021, pp. 7710--7720.

\bibitem{face_detection}
Z.~Liu, J.~Yang, M.~Lin, K.~K.~F. Lai, S.~Yanushkevich, O.~Yadid-Pecht, Wdr face: The first database for studying face detection in wide dynamic range (2021).
\newblock \href {http://arxiv.org/abs/2101.03826} {\path{arXiv:2101.03826}}.

\bibitem{Eil}
G.~Eilertsen, J.~Kronander, G.~Denes, R.~Mantiuk, J.~Unger, Hdr image reconstruction from a single exposure using deep cnns, ACM Transactions on Graphics 36~(6) (nov 2017).
\newblock \href {https://doi.org/10.1145/3130800.3130816} {\path{doi:10.1145/3130800.3130816}}.

\bibitem{Tel_dataset}
S.~Tel, Z.~Wu, Y.~Zhang, B.~Heyrman, C.~Demonceaux, R.~Timofte, D.~Ginhac, Alignment-free hdr deghosting with semantics consistent transformer, in: 2023 IEEE/CVF International Conference on Computer Vision (ICCV), 2023, pp. 12790--12799.
\newblock \href {https://doi.org/10.1109/ICCV51070.2023.01179} {\path{doi:10.1109/ICCV51070.2023.01179}}.

\bibitem{hyperspectral}
X.~Zhang, H.~Zhao, Hyperspectral-cube-based mobile face recognition: A comprehensive review, Information Fusion 74 (2021) 132--150.
\newblock \href {https://doi.org/10.1016/j.inffus.2021.04.003} {\path{doi:10.1016/j.inffus.2021.04.003}}.

\bibitem{DFNet}
J.~Peng, H.~Zhao, Z.~Hu, Dynamic fusion network for rgbt tracking, IEEE Transactions on Intelligent Transportation Systems 24~(4) (2023) 3822--3832.
\newblock \href {https://doi.org/10.1109/TITS.2022.3229830} {\path{doi:10.1109/TITS.2022.3229830}}.

\bibitem{HDR2}
J.~Liu, G.~Wu, J.~Luan, Z.~Jiang, R.~Liu, X.~Fan, \href{https://www.sciencedirect.com/science/article/pii/S1566253523000672}{Holoco: Holistic and local contrastive learning network for multi-exposure image fusion}, Information Fusion 95 (2023) 237--249.
\newblock \href {https://doi.org/https://doi.org/10.1016/j.inffus.2023.02.027} {\path{doi:https://doi.org/10.1016/j.inffus.2023.02.027}}.
\newline\urlprefix\url{https://www.sciencedirect.com/science/article/pii/S1566253523000672}

\bibitem{HDR3}
H.~Zhang, J.~Ma, \href{https://www.sciencedirect.com/science/article/pii/S1566253523000787}{Iid-mef: A multi-exposure fusion network based on intrinsic image decomposition}, Information Fusion 95 (2023) 326--340.
\newblock \href {https://doi.org/https://doi.org/10.1016/j.inffus.2023.02.031} {\path{doi:https://doi.org/10.1016/j.inffus.2023.02.031}}.
\newline\urlprefix\url{https://www.sciencedirect.com/science/article/pii/S1566253523000787}

\bibitem{HDR4}
J.~Zhang, Y.~Luo, J.~Huang, Y.~Liu, J.~Ma, \href{https://www.sciencedirect.com/science/article/pii/S1566253523002117}{Multi-exposure image fusion via perception enhanced structural patch decomposition}, Information Fusion 99 (2023) 101895.
\newblock \href {https://doi.org/https://doi.org/10.1016/j.inffus.2023.101895} {\path{doi:https://doi.org/10.1016/j.inffus.2023.101895}}.
\newline\urlprefix\url{https://www.sciencedirect.com/science/article/pii/S1566253523002117}

\bibitem{Debevec+1997}
P.~E. Debevec, J.~Malik, Recovering high dynamic range radiance maps from photographs, in: Proceedings of the 24th Annual Conference on Computer Graphics and Interactive Techniques, SIGGRAPH '97, ACM Press/Addison-Wesley Publishing Co., USA, 1997, p. 369–378.
\newblock \href {https://doi.org/10.1145/258734.258884} {\path{doi:10.1145/258734.258884}}.

\bibitem{Lecouat+2022}
B.~Lecouat, T.~Eboli, J.~Ponce, J.~Mairal, High dynamic range and super-resolution from raw image bursts, ACM Transactions on Graphics 41~(4) (jul 2022).
\newblock \href {https://doi.org/10.1145/3528223.3530180} {\path{doi:10.1145/3528223.3530180}}.

\bibitem{SiameseTracking}
J.~Peng, H.~Zhao, Z.~Hu, Y.~Zhuang, B.~Wang, Siamese infrared and visible light fusion network for rgb-t tracking, International Journal of Machine Learning and Cybernetics 14~(9) (2023) 3281--3293.
\newblock \href {https://doi.org/10.1007/s13042-023-01833-6} {\path{doi:10.1007/s13042-023-01833-6}}.

\bibitem{RGBTLowLight}
C.~Yu, S.~Li, W.~Feng, T.~Zheng, S.~Liu, Saca-fusion: a low-light fusion architecture of infrared and visible images based on self- and cross-attention, The Visual Computer 40~(5) (2024) 3347--3356.
\newblock \href {https://doi.org/10.1007/s00371-023-03037-z} {\path{doi:10.1007/s00371-023-03037-z}}.

\bibitem{Kalantari+2017}
N.~K. Kalantari, R.~Ramamoorthi, Deep high dynamic range imaging of dynamic scenes, ACM Transactions on Graphics 36~(4) (jul 2017).
\newblock \href {https://doi.org/10.1145/3072959.3073609} {\path{doi:10.1145/3072959.3073609}}.

\bibitem{Mobile_HDR}
S.~Liu, X.~Zhang, L.~Sun, Z.~Liang, H.~Zeng, L.~Zhang, Joint hdr denoising and fusion: A real-world mobile hdr image dataset, in: 2023 IEEE/CVF Conference on Computer Vision and Pattern Recognition (CVPR), IEEE Computer Society, Los Alamitos, CA, USA, 2023, pp. 13966--13975.
\newblock \href {https://doi.org/10.1109/CVPR52729.2023.01342} {\path{doi:10.1109/CVPR52729.2023.01342}}.

\bibitem{LasHeR}
C.~Li, W.~Xue, Y.~Jia, Z.~Qu, B.~Luo, J.~Tang, D.~Sun, Lasher: A large-scale high-diversity benchmark for rgbt tracking, IEEE Transactions on Image Processing 31 (2022) 392--404.
\newblock \href {https://doi.org/10.1109/TIP.2021.3130533} {\path{doi:10.1109/TIP.2021.3130533}}.

\bibitem{FLIR}
S.~Banerjee, Teledyne flir adas thermal dataset v2, \url{http://www.kaggle.com/datasets/samdazel/teledyne-flir-adas-thermal-dataset-v2} (2024).

\bibitem{GTOT}
C.~Li, H.~Cheng, S.~Hu, X.~Liu, J.~Tang, L.~Lin, Learning collaborative sparse representation for grayscale-thermal tracking, IEEE Transactions on Image Processing 25~(12) (2016) 5743--5756.
\newblock \href {https://doi.org/10.1109/TIP.2016.2614135} {\path{doi:10.1109/TIP.2016.2614135}}.

\bibitem{RGBT234}
C.~Li, X.~Liang, Y.~Lu, N.~Zhao, J.~Tang, Rgb-t object tracking: Benchmark and baseline, Pattern Recognition 96 (2019) 106977.
\newblock \href {https://doi.org/10.1016/j.patcog.2019.106977} {\path{doi:10.1016/j.patcog.2019.106977}}.

\bibitem{LLVIP}
X.~Jia, C.~Zhu, M.~Li, W.~Tang, W.~Zhou, Llvip: A visible-infrared paired dataset for low-light vision, in: 2021 IEEE/CVF International Conference on Computer Vision Workshops (ICCVW), 2021, pp. 3489--3497.
\newblock \href {https://doi.org/10.1109/ICCVW54120.2021.00389} {\path{doi:10.1109/ICCVW54120.2021.00389}}.

\bibitem{NTIRE21}
E.~P{\'{e}}rez{-}Pellitero, S.~Catley{-}Chandar, A.~Leonardis, R.~Timofte, {NTIRE} 2021 challenge on high dynamic range imaging: Dataset, methods and results, in: {IEEE} Conference on Computer Vision and Pattern Recognition Workshops, {CVPR} Workshops 2021, virtual, June 19-25, 2021, Computer Vision Foundation / {IEEE}, 2021, pp. 691--700.

\bibitem{yu2021luminance}
H.~Yu, W.~Liu, C.~Long, B.~Dong, Q.~Zou, C.~Xiao, Luminance attentive networks for hdr image and panorama reconstruction, Computer Graphics Forum 40~(7) (2021) 181--192.

\bibitem{Zhang+2021}
Y.~Zhang, T.~O. Aydin, Deep {HDR} estimation with generative detail reconstruction, Comput. Graph. Forum 40~(2) (2021) 179--190.
\newblock \href {https://doi.org/10.1111/cgf.142624} {\path{doi:10.1111/cgf.142624}}.

\bibitem{DeepRecursiveHDRI}
S.~Lee, G.~H. An, S.-J. Kang, Deep recursive hdri: Inverse tone mapping using generative adversarial networks, in: V.~Ferrari, M.~Hebert, C.~Sminchisescu, Y.~Weiss (Eds.), Computer Vision -- ECCV 2018, Springer International Publishing, Cham, 2018, pp. 613--628.

\bibitem{Zhang_2023_CVPR}
N.~Zhang, Y.~Ye, Y.~Zhao, R.~Wang, Revisiting the stack-based inverse tone mapping, in: Proceedings of the IEEE/CVF Conference on Computer Vision and Pattern Recognition (CVPR), 2023, pp. 9162--9171.

\bibitem{Prabhakar_dataset}
K.~R. Prabhakar, S.~Agrawal, D.~K. Singh, B.~Ashwath, R.~V. Babu, Towards practical and efficient high-resolution hdr deghosting with cnn, in: A.~Vedaldi, H.~Bischof, T.~Brox, J.-M. Frahm (Eds.), Computer Vision -- ECCV 2020, Springer International Publishing, Cham, 2020, pp. 497--513.

\bibitem{wang2022glowgan}
C.~Wang, A.~Serrano, X.~Pan, B.~Chen, H.-P. Seidel, C.~Theobalt, K.~Myszkowski, T.~Leimkuehler, Glowgan: Unsupervised learning of hdr images from ldr images in the wild, in: Proceedings of the IEEE/CVF International Conference on Computer Vision (ICCV), IEEE, 2023.

\bibitem{Banterle+2024}
F.~Banterle, D.~Marnerides, T.~Bashford-rogers, K.~Debattista, Self-supervised high dynamic range imaging: What can be learned from a single 8-bit video?, ACM Transactions on Graphics 43~(2) (mar 2024).
\newblock \href {https://doi.org/10.1145/3648570} {\path{doi:10.1145/3648570}}.

\bibitem{Corneanu+2016}
C.~A. Corneanu, M.~O. Simón, J.~F. Cohn, S.~E. Guerrero, Survey on rgb, 3d, thermal, and multimodal approaches for facial expression recognition: History, trends, and affect-related applications, IEEE Transactions on Pattern Analysis and Machine Intelligence 38~(8) (2016) 1548--1568.
\newblock \href {https://doi.org/10.1109/TPAMI.2016.2515606} {\path{doi:10.1109/TPAMI.2016.2515606}}.

\bibitem{RGBTFusion}
J.~Ma, Y.~Ma, C.~Li, Infrared and visible image fusion methods and applications: A survey, Information Fusion 45 (2019) 153--178.
\newblock \href {https://doi.org/10.1016/j.inffus.2018.02.004} {\path{doi:10.1016/j.inffus.2018.02.004}}.

\bibitem{RGBTTracking}
X.~Zhang, P.~Ye, H.~Leung, K.~Gong, G.~Xiao, Object fusion tracking based on visible and infrared images: A comprehensive review, Information Fusion 63 (2020) 166--187.
\newblock \href {https://doi.org/10.1016/j.inffus.2020.05.002} {\path{doi:10.1016/j.inffus.2020.05.002}}.

\bibitem{RGBTdetection}
M.~Yuan, X.~Shi, N.~Wang, Y.~Wang, X.~Wei, Improving rgb-infrared object detection with cascade alignment-guided transformer, Information Fusion 105 (2024) 102246.
\newblock \href {https://doi.org/10.1016/j.inffus.2024.102246} {\path{doi:10.1016/j.inffus.2024.102246}}.

\bibitem{RGBTColor}
S.~Yin, L.~Cao, Y.~Ling, G.~Jin, One color contrast enhanced infrared and visible image fusion method, Infrared Physics \& Technology 53~(2) (2010) 146--150.
\newblock \href {https://doi.org/10.1016/j.infrared.2009.10.007} {\path{doi:10.1016/j.infrared.2009.10.007}}.

\bibitem{Pixel2}
X.~Zhang, Y.~Demiris, Visible and infrared image fusion using deep learning, IEEE Transactions on Pattern Analysis and Machine Intelligence 45~(8) (2023) 10535--10554.
\newblock \href {https://doi.org/10.1109/TPAMI.2023.3261282} {\path{doi:10.1109/TPAMI.2023.3261282}}.

\bibitem{Decision}
P.~Zhang, J.~Zhao, C.~Bo, D.~Wang, H.~Lu, X.~Yang, Jointly modeling motion and appearance cues for robust rgb-t tracking, IEEE Transactions on Image Processing 30 (2021) 3335--3347.
\newblock \href {https://doi.org/10.1109/TIP.2021.3060862} {\path{doi:10.1109/TIP.2021.3060862}}.

\bibitem{collapse}
I.~Shumailov, Z.~Shumaylov, Y.~Zhao, N.~Papernot, R.~Anderson, Y.~Gal, Ai models collapse when trained on recursively generated data, Nature 631~(8022) (2024) 755--759.
\newblock \href {https://doi.org/10.1038/s41586-024-07566-y} {\path{doi:10.1038/s41586-024-07566-y}}.

\bibitem{EDSR}
B.~Lim, S.~Son, H.~Kim, S.~Nah, K.~M. Lee, Enhanced deep residual networks for single image super-resolution, in: The IEEE Conference on Computer Vision and Pattern Recognition (CVPR) Workshops, 2017.

\bibitem{ExpandNet}
D.~Marnerides, T.~Bashford-Rogers, J.~Hatchett, K.~Debattista, Expandnet: A deep convolutional neural network for high dynamic range expansion from low dynamic range content, Computer Graphics Forum 37~(2) (2018) 37--49.
\newblock \href {https://doi.org/10.1111/cgf.13340} {\path{doi:10.1111/cgf.13340}}.

\bibitem{GAN}
I.~Goodfellow, J.~Pouget-Abadie, M.~Mirza, B.~Xu, D.~Warde-Farley, S.~Ozair, A.~Courville, Y.~Bengio, Generative adversarial networks, Commun. ACM 63~(11) (2020) 139--144.
\newblock \href {https://doi.org/10.1145/3422622} {\path{doi:10.1145/3422622}}.

\bibitem{End}
Y.~Endo, Y.~Kanamori, J.~Mitani, Deep reverse tone mapping, ACM Transactions on Graphics 36~(6) (nov 2017).
\newblock \href {https://doi.org/10.1145/3130800.3130834} {\path{doi:10.1145/3130800.3130834}}.

\bibitem{HDRTVNet}
X.~Chen, Z.~Zhang, J.~S. Ren, L.~Tian, Y.~Qiao, C.~Dong, A new journey from sdrtv to hdrtv, in: 2021 IEEE/CVF International Conference on Computer Vision (ICCV), 2021, pp. 4480--4489.
\newblock \href {https://doi.org/10.1109/ICCV48922.2021.00446} {\path{doi:10.1109/ICCV48922.2021.00446}}.

\bibitem{LaNet}
H.~Yu, W.~Liu, C.~Long, B.~Dong, Q.~Zou, C.~Xiao, Luminance attentive networks for hdr image and panorama reconstruction, Computer Graphics Forum 40~(7) (2021) 181--192.
\newblock \href {https://doi.org/10.1111/cgf.14412} {\path{doi:10.1111/cgf.14412}}.

\bibitem{SAN}
M.~S. Santos, T.~I. Ren, N.~K. Kalantari, Single image hdr reconstruction using a cnn with masked features and perceptual loss, ACM Transactions on Graphics 39~(4) (aug 2020).
\newblock \href {https://doi.org/10.1145/3386569.3392403} {\path{doi:10.1145/3386569.3392403}}.

\bibitem{ICTCPNet}
P.~Huang, G.~Cao, F.~Zhou, G.~Qiu, Video inverse tone mapping network with luma and chroma mapping, in: Proceedings of the 31st ACM International Conference on Multimedia, MM '23, Association for Computing Machinery, New York, NY, USA, 2023, pp. 1383--1391.
\newblock \href {https://doi.org/10.1145/3581783.3612199} {\path{doi:10.1145/3581783.3612199}}.

\bibitem{HDRUNet}
X.~Chen, Y.~Liu, Z.~Zhang, Y.~Qiao, C.~Dong, Hdrunet: Single image hdr reconstruction with denoising and dequantization, in: 2021 IEEE/CVF Conference on Computer Vision and Pattern Recognition Workshops (CVPRW), 2021, pp. 354--363.
\newblock \href {https://doi.org/10.1109/CVPRW53098.2021.00045} {\path{doi:10.1109/CVPRW53098.2021.00045}}.

\bibitem{PU21}
R.~K. Mantiuk, M.~Azimi, Pu21: A novel perceptually uniform encoding for adapting existing quality metrics for hdr, in: 2021 Picture Coding Symposium (PCS), 2021, pp. 1--5.
\newblock \href {https://doi.org/10.1109/PCS50896.2021.9477471} {\path{doi:10.1109/PCS50896.2021.9477471}}.

\bibitem{UNet}
O.~Ronneberger, P.~Fischer, T.~Brox, U-net: Convolutional networks for biomedical image segmentation, in: Medical Image Computing and Computer Assisted Intervention, PT III, Vol. 9351, 2015, pp. 234--241.
\newblock \href {https://doi.org/10.1007/978-3-319-24574-4\_28} {\path{doi:10.1007/978-3-319-24574-4\_28}}.

\bibitem{Durand}
F.~Durand, J.~Dorsey, Fast bilateral filtering for the display of high-dynamic-range images, ACM Transactions on Graphics 21~(3) (2002) 257–266.
\newblock \href {https://doi.org/10.1145/566654.566574} {\path{doi:10.1145/566654.566574}}.

\end{thebibliography}

\end{document}